\documentclass[]{fairmeta}
\usepackage{booktabs}
\usepackage{amsmath}
\usepackage{amssymb}
\title{Teaching Large Language Models to Reason with Reinforcement Learning}

\author[1,2,*]{Alex Havrilla}
\author[4]{Yuqing Du}
\author[1]{Sharath Chandra Raparthy}
\author[1]{Christoforos Nalmpantis}
\author[1]{Jane Dwivedi-Yu}
\author[3]{Maksym Zhuravinskyi}
\author[1,**]{Eric Hambro}
\author[1]{Sainbayar Sukhbaatar}
\author[1]{Roberta Raileanu}

\affiliation[1]{Meta}
\affiliation[2]{Georgia Institute of Technology}
\affiliation[3]{StabilityAI}
\affiliation[4]{UC Berkeley}

\contribution[*]{Work done during Meta internship}
\contribution[**]{Work done while at Meta}

\abstract{Reinforcement Learning from Human Feedback (\textbf{RLHF}) has emerged as a dominant approach for aligning LLM outputs with human preferences. Inspired by the success of RLHF, we study the performance of multiple algorithms that learn from feedback (Expert Iteration, Proximal Policy Optimization (\textbf{PPO}), Return-Conditioned RL) on improving LLM reasoning capabilities. We investigate both sparse and dense rewards provided to the LLM both heuristically and via a learned reward model. We additionally start from multiple model sizes and initializations both with and without supervised fine-tuning (\textbf{SFT}) data.  Overall, we find all algorithms perform comparably, with Expert Iteration performing best in most cases. Surprisingly, we find the sample complexity of Expert Iteration is similar to that of PPO, requiring at most on the order of $10^6$ samples to converge from a pretrained checkpoint. We investigate why this is the case, concluding that during RL training models fail to explore significantly beyond solutions already produced by SFT models. Additionally, we discuss a trade off between maj@1 and pass@96 metric performance during SFT training and how conversely RL training improves both simultaneously. We then conclude by discussing the implications of our findings for RLHF and the future role of RL in LLM fine-tuning.}

\date{\today}
\correspondence{Alex Havrilla at \email{ahavrilla3@gatech.edu}}

\begin{document}

\maketitle

\section{Introduction}
The reasoning abilities of large language models (\textbf{LLMs}) are rapidly improving as measured by their performance on numerous math, science and code benchmarks \citep{Cobbe2021TrainingVT, Hendrycks2021MeasuringMP, Sawada2023ARBAR, Liang2022HolisticEO, Srivastava2022BeyondTI, Rein2023GPQAAG, Mialon2023GAIAAB, Chollet2019OnTM, Mishra2022LilaAU, hendrycks2021measuring, Austin2021ProgramSW, patel2021nlp, eval-harness}. Simultaneously, Reinforcement Learning from Human Feedback (RLHF) \citep{Bai2022ConstitutionalAH, ziegler2019fine, Ouyang2022TrainingLM} and instruction fine-tuning \citep{Wei2021FinetunedLM, Mishra2021CrossTaskGV} have made significant progress in aligning LLMs with human preferences. Improvements in model instructability have further increased apparent model capability by making complex behaviors more accessible via instruction prompting. This has led to a number of increasingly sophisticated prompting strategies augmenting LLM reasoning capabilities such as Chain-of-Thought \citep{Wei2022ChainOT} or Tree-of-Thoughts \citep{Yao2023TreeOT}.


Previous work in reinforcement learning (RL) such as AlphaGo \citep{Silver2017MasteringCA}, AlphaStar \citep{Vinyals2019GrandmasterLI}, and OpenAI Dota 2 \citep{Berner2019Dota2W} demonstrate that RL techniques can be used to train neural networks capable of sophisticated planning and reasoning in game environments. Cicero \citep{cicero} in particular succeeds in combining an RL trained planning agent with a dialogue fine-tuned LLM to achieve nearly super-human performance in the board game Diplomacy. Given these previous successes and the inherent interactive nature of problem solving, applying RL to LLM reasoning seems a natural next step. In this paper, we study how ideas from RL can be used to improve the reasoning capabilities of LLMs across a variety of reward schemes and model initializations.

We begin by comparing the performance of different RL algorithms on reasoning tasks $\tau$ defined as a distribution of question answer tuples $(Q, A)$. The task $\tau$ can be extended to define a \textit{Markov Decision Process} (\textbf{MDP}) 4-tuple $(\mathcal{S}, \mathcal{A}, P_a, R_a)$ where tokens serve as both actions and accumulated state with deterministic dynamics. By default we use a sparse reward of $+1$ if the final answer is correct but also experiment with dense rewards matching intermediate steps in a reference solution and rewards synthetically generated using a reward model. We evaluate models with 7B and 13B parameters both starting from supervised fine-tuned (SFT) checkpoints and pre-trained checkpoints. We report four metrics assessing model performance on a task specific test set: 1) maj@1 score computed by greedily sampling once per question, 2) maj@96 score computed by sampling K = 96 times per question and uniformly voting on the final answer, 3) rerank@96 score computed by sampling K = 96 times and choosing the final answer using an Outcome-Based Reward Model (\textbf{ORM}), and 4) pass@96 score computed by sampling the model K = 96 times and taking the best result according to the ground truth answer.

We find that overall the simplest method, Expert Iteration (\textbf{EI}) \citep{Anthony2017ThinkingFA}, performs best across all metrics for most reward setups and model initializations. Surprisingly, EI is nearly as sample efficient as more sophisticated algorithms like Proximal Policy Optimization (\textbf{PPO}), both requiring only a few thousand samples to converge even when initialized from a pretrained checkpoint. We also observe the gap between pretrained model performance and SFT model performance significantly shrinks ($<$ 10\% gap on GSM8K) after RL fine-tuning, with larger models having a smaller gap. Additionally, previous work identified a tradeoff between test time maj@1 performance and pass@96 performance during supervised fine-tuning \citep{Cobbe2021TrainingVT}, with continued training increasing maj@1 score at the expense of pass@96 score. We identify the limited diversity of the dataset as a core reason for this. We show that RL fine-tuning can improve both metrics simultaneously due to the fact that RL generates its own data during training, resulting in a more diverse set of examples to learn from. 

We then discuss why EI and return conditioned RL are competitive with PPO, suggesting two principal factors. Firstly, the reasoning tasks we consider have entirely deterministic dynamics: a setting in which direct behavior cloning and return conditioned RL is known to do well \citep{Brandfonbrener2022WhenDR}. In contrast, PPO often succeeds in environments with a high degree of stochasticity \citep{Bhargava2023SequenceMI}. Second, we identify a lack of sophisticated exploration carried out by models during RL fine-tuning. This limitation significantly impacts any performance or sample complexity advantages PPO may have when fine-tuning the pretrained model. We come to this conclusion from a number of observations, noting in particular quickly saturating pass@96 scores early in RL training. We conclude with a discussion of the impacts of our observations on RLHF and the future of LLM fine-tuning via RL.




In summary we make the following contributions:

\begin{itemize}
    \item A comprehensive study of PPO fine-tuning of LLMs on reasoning tasks using different types of rewards, model sizes and initializations.
    \item A comparison to expert iteration and return-conditioned RL from which we find expert iteration reliably attains the best performance and competitive sample complexity across the board.
    \item A discussion of the implications of our findings for RLHF and the future of RL fine-tuning for LLMs, identifying exploration as a major limiting factor.
\end{itemize}

\section{Related Work}

\textbf{LLM Reasoning:} 
State-of-the-art large language models \citep{OpenAI2023GPT4TR, touvron2023llama, Bai2022ConstitutionalAH, Chowdhery2022PaLMSL} demonstrate increasingly impressive abilties on hard reasoning tasks as studied by a wide range of math, science, and code benchmarks \citep{Cobbe2021TrainingVT, Hendrycks2021MeasuringMP, Sawada2023ARBAR, Liang2022HolisticEO, Srivastava2022BeyondTI, Rein2023GPQAAG, Mialon2023GAIAAB, Chollet2019OnTM, Mishra2022LilaAU, hendrycks2021measuring, Austin2021ProgramSW, patel2021nlp, eval-harness}. \textit{Chain of thought} (\textbf{CoT}) \citep{Wei2022ChainOT} and related techniques \citep{Chen2022ProgramOT, Yao2023TreeOT, Besta2023GraphOT} have emerged as dominant methods siginficantly boosting LLM performance on these types of tasks. CoT methods allow LLMs to defer giving their final answer by first generating a "chain of thought" involving intermediate computations needed to correctly solve the problem. 

Another line of work combines base LLM reasoning capabilities with planning and search algorithms to further boost performance on a wide range of tasks \citep{Yao2023TreeOT, Besta2023GraphOT, Ye2022NeuralSP, Yao2022ReActSR, Dohan2022LanguageMC}. Tree of thought \citep{Yao2023TreeOT} for example combines LLMs with a breadth first search algorithm, relying on the LLM to both propose actions and evaluate state. Other works combine LLMs with tools \citep{Schick2023ToolformerLM, Qin2023ToolLLMFL, Zhou2023SolvingCM} 
further boosting reasoning capability. Combining GPT-4 with a python code interpreter for generation and self-verification achieves an impressive 84\% on the hard MATH benchmark \citep{hendrycks2021measuring, Zhou2023SolvingCM}.


Other works focus on LLMs for mathematical reasoning in natural language \citep{Cobbe2021TrainingVT, Lewkowycz2022SolvingQR, Azerbayev2023LlemmaAO, Lightman2023LetsVS, patel2021nlp, Zhu_2023, Rafailov2023DirectPO}. Particularly relevant to our study is \citet{Cobbe2021TrainingVT} which fine-tunes GPT-3 on supervised math word problem (\textbf{MWP}) reasoning traces. In addition they train solution verifiers called Outcome Based Reward Models (\textbf{ORMs}) which predict the probability of correctly solving a question $Q$ giving a prefix of intermediate steps $P_i = (S_1, ..., S_i)$ i.e. $p(is\_correct(A) | Q, P_i)$ where $A$ is a solution with prefix $P_i$. Process based reward models (\textbf{PRMs}) \citep{Uesato2022SolvingMW, Lightman2023LetsVS} can also be trained to instead look at the step-level accuracy of solutions. More recent work \citep{Luo2023WizardMathEM} utlizies a PRM distilled from GPT-4 feedback as a reward signal during PPO.


\textbf{RL for LLM fine-tuning: } Reinforcement Learning from Human Feedback (RLHF) is perhaps the most well-known application of RL techniques for fine-tuning LLMs. RLHF \citep{christiano2017deep, ziegler2019fine, Stiennon2020LearningTS, Ouyang2022TrainingLM, Bai2022ConstitutionalAH, Glaese2022ImprovingAO, peng2021inferring, Ramamurthy2022IsRL} most often works by training a \textit{reward model} to capture human preferences over a task $\tau$. The reward model is then used to score LLM responses to prompts from the task after which policy improvement is performed. PPO is most often used \citep{Ouyang2022TrainingLM, Bai2022ConstitutionalAH} but several recent works including ReST \citep{Gulcehre2023ReinforcedS}, Reward-Ranked Fine-tuning \citep{Dong2023RAFTRR}, and AlpacaFarm \citep{Dubois2023AlpacaFarmAS} all demonstrate simply fine-tuning on high return responses with the standard cross-entropy loss can attain comparable performance. We broadly refer to this class of algorithms as Expert Iteration.

A large body of work studying RL for LLM fine-tuning also exists outside of the RLHF sphere. Work on text games \citep{Yao2020KeepCA, ammanabrolu-riedl-2019-playing} and other interactive textual environments \citep{Zhou2023DialogueSE, Carta2023GroundingLL} seek to ground LLMs via interaction and RL. RL has also been applied to improving model performance on controllable generation and question answering tasks \citep{Lu2022QuarkCT, Liu2022RainierRK}. Various forms of expert iteration have also been applied to improve LLM reasoning capabilities \citep{Huang2022LargeLM, Yuan2023ScalingRO, zelikman2022star, Uesato2022SolvingMW}. For example  ``Scaling Relationship on Learning Mathematical Reasoning with Large Language Models'' \citep{Yuan2023ScalingRO} applies a single round of expert iteration across multiple model sizes on GSM8K. They observe sizeable gains in all metrics for smaller models, with gains diminishing for larger models. A related body of work studies RL for code generation \citep{Le2022CodeRLMC, Shen2023PanGuCoder2BL, rozière2023code}. \citet{Shen2023PanGuCoder2BL} in particular reports a huge increase in StarCoder's \citep{Li2023StarCoderMT} maj@1 performance after a single round of expert iteration, jumping from $\sim$30\% to $\sim$60\%.

Despite all the above work, it remains unclear exactly what factors account for the biggest impact during RL fine-tuning due to wide variance in tasks, pretraining data, supervised fine-tuning data, RL algorithm used, and the reward source. Our work conducts a thorough analysis of all these factors to understand exactly how different algorithms compare when applied to improving LLM reasoning capability. As a result we are able to identify key bottlenecks to further LLM improvement via RL and provide a discussion on promising future directions.

\section{Methods}
\label{sec:methods}

\textbf{Reasoning as an RL problem} 

We study the performance and sample complexity requirements for various RL algorithms when fine-tuning LLMs on reasoning tasks. We consider Expert Iteration (EI) \citep{Anthony2017ThinkingFA}, Proximal Policy Optimization (PPO) \citep{Schulman2017ProximalPO}, and Return-Conditioned RL (RCRL) \citep{Brandfonbrener2022WhenDR} as representative algorithms from the RL literature. 
In general, the goal of all RL algorithms is to maximize the expected future return $\mathbb{E}_{A \sim \pi(Q), (Q, \cdot) \in \tau} R(A)$ of a student policy $\pi$ on task $\tau$. We call the highest return policy the \textit{optimal policy} $\pi^*$. Each of our chosen algorithms goes about finding $\pi^*$ in a different way.

\textbf{PPO} is an example of an \textit{online} RL algorithm. Online algorithms engage in both an exploration phase and a policy improvement phase which updates $\pi_\theta$ using data generated during the exploration phase. PPO is also an \textit{on-policy} algorithm which samples model rollouts during exploration from the student policy $\pi_\theta$ being trained. During policy improvement, the student $\pi_\theta$ updates its parameters via gradient descent by directly maximizing for reward with the objective 

\begin{align*}
    J(\theta) = \mathbb{E}_t \left[ min(\frac{\pi(a_t | s_t)}{\pi_\textup{old}(a_t | s_t)}  \hat{A}_t, clip(1-\epsilon, 1+\epsilon, \frac{\pi(a_t | s_t)}{\pi_\textup{old}(a_t | s_t)})\hat{A}_t) \right]
\end{align*} where $\hat{A}_t$ estimates the \textit{advantage} i.e. difference between $Q(s,a)$ (the expected return after taking action $a$ at state $s$) and value $V(s)$ (the expected return at state $s$). 


In practice, for PPO we sample 1024 rollouts at a time with a temperature of 0.7 and $N=4$ rollouts per question. Training is then run on these samples for $K=4$ PPO epochs with a batch size of 256. Additionally, we train using LoRA \citep{Hu2021LoRALA} with $r=128$. Training is run for 4000 gradient steps. The best checkpoint is then selected via performance on a validation set. 

\textbf{Expert iteration} is also online but more off-policy than PPO. An initial expert policy approximation $\hat{\pi}^*_0$ is sampled on the entire train set $K$ times per question before any policy improvement. The $\hat{\pi}^*_0$ is often constructed using repeated sampling from an initial policy $\pi_0$. For example, AlphaZero \citep{Silver2017MasteringCA} and subsequent work \citep{Schick2023ToolformerLM} combine $\pi_0$ with Monte Carlo Tree Search. Sampling $\hat{\pi}^*_0$ constructs an initial set of rollouts $D_1$ which are then distilled back into a policy $\pi_1$ via a standard cross-entropy loss: $\sum_{\tau \in D} \sum_{t=1}^H -log(\pi_\theta(a_t | s_t))$. This process can be repeated to construct policy $\pi_i$ fine-tuned on dataset $D_i = R_i \cup D_{i-1}$ where $R_i$ corresponds to exploration done by $\pi_{i-1}$.

In our setting we construct an approximation to the optimal policy $\hat{\pi}^*$ by rejection sampling our student policy $\pi_\theta$. After generating $K$ samples $S_1,..., S_K$ on a question $Q$ we construct $D_1$ by filtering all $(Q, S_i)$ pairs with return below a threshold $T$. De-duplication is then performed on the remaining samples.

In practice, during the expert iteration exploration phase we sample each question in the train set $K=96$ times with temperature $T = 1.0$. To construct the training set we filter out incorrect solutions and duplicates. Importantly, fine-tuning is then done from the pretrained base model with the same hyperparameters as SFT. This is repeated until performance on a validation set saturates.

\textbf{Return Conditioned RL} Return conditioned RL algorithms seek to train policies conditioned on both the current state $s$ and desired return $R$ when sampling an action. This is motivated by a desire to learn return conditionable policies which can change depending on the desired return. Best performance can then be sampled by conditioning on the highest possible return. 

We consider an offline version of this class of algorithms similar to a decision transformer \citep{Chen2021DecisionTR}. A training dataset $D$ is constructed by generating state, action, return $\tau = ((s_t, a_t, g_t))_{t=1}^H$ trajectories. Training is done by predicting the action given state and return: $\sum_{\tau \in D} \sum_{t=1}^H -log(\pi_\theta(a_t | s_t, g_t))$. In practice we construct $D$ by sampling solutions $S = (S_1, ..., S_L)$, where each $S_i$ is an intermediate step, from our best EI trained policy $\pi_\textup{EI}$ given a question $Q$. We generate return labels for each step $S_i$ by sampling $\pi_\textup{EI}$ K many times from $P_i = (S_1, ..., S_i)$. This results in binary labels $l_1,.., l_K$ evaluating the correctness of the generated final answers. $S_i$ is then labeled as ``[GOOD]'' if the average return $\frac{1}{K}\sum_{k=1}^K l_k \geq T$ and otherwise is labeled as ``[BAD]''. Typically we set $T = 0.5$. We then filter the dataset to ensure a balanced number of correct and incorrect solutions. See Section \ref{sec:sorm} in the appendix for more details about the step-label generating process.

\textbf{Outcome Based Reward Modeling} Multiple works \citep{Cobbe2021TrainingVT, Uesato2022SolvingMW} train Outcome Based Reward models \textbf{ORMs} as \textit{verifiers} of candidate solutions to word problems. The ORM can then be used to rerank multiple candidate solutions generated by a student model, significantly boosting performance. Training data for the ORM is generated using a student policy $\pi$ by sampling $K$ solutions per question $Q$ in the task dataset. The ORM is trained as a classifier by predicting the probability of reaching the correct final answer $p(\texttt{is\_correct(A)} | Q, P_i)$ from an intermediate sequence of steps $P_i = (S_1,..., S_i)$, $P_i \subseteq A = (S_1,...,S_L)$.

\section{Experiments}
\label{sec:experiments}


We conduct our evaluations on GSM8K and SVAMP \citep{patel2021nlp}: two math word problem benchmarks. In addition on GSM8K we consider two data regimes: first with SFT data and then without SFT data. We evaluate all models using greedy sampling (maj@1) accuracy as well majority vote at 96 samples (maj@96), ORM based reranking at 96 samples (rerank@96), and best of 96 sample (pass@96) accuracy. Unless otherwise specified, test-time sampling is done greedily for maj@1 and with a temperature of 0.7 otherwise. We sample the RCRL models one step/line at a time, conditioning on the ``[GOOD]'' token. We note while the notion of a ``step'' is not clearly defined in general, in our case we can simply regard each step as ending with a sentence or newline. All experiments are done using instruction-tuned Llama-2 7B and Llama-2 13B models.

\begin{table}[t]
    \centering
    \begin{tabular}{@{}lrrrrrrrr@{}}
        \toprule
        & \multicolumn{2}{c}{maj@1} & \multicolumn{2}{c}{maj@96} & \multicolumn{2}{c}{rerank@96$^\dagger$} & \multicolumn{2}{c}{pass@96} \\
        \cmidrule(lr){2-3} \cmidrule(lr){4-5} \cmidrule(lr){6-7} \cmidrule(lr){8-9}
        & 7B & 13B & 7B & 13B & 7B & 13B & 7B & 13B \\
        \midrule
        SFT & 0.41 & 0.48 & 0.47 & 0.53 & 0.54 & 0.68 & 0.72 & 0.84 \\
        EI$_n$ & \textbf{0.48} & \textbf{0.53} & \textbf{0.55} & \textbf{0.59} & 0.64 & \textbf{0.71} & 0.8 & \textbf{0.88} \\ 
         ORM EI$_n$ & \textbf{0.48} & \textbf{0.53} & 0.54 & 0.58 & \textbf{0.65} & \textbf{0.71} & \textbf{0.81} & 0.87 \\
         ORM RCRL & 0.45 & 0.51 & 0.5 & 0.56 & 0.54 & 0.69 & 0.73 & 0.83 \\
         Sparse PPO & 0.44 & 0.51 & 0.49 & 0.55 & 0.58 & 0.67 & 0.77 & 0.85 \\
         Dense PPO & 0.43 & 0.50 & 0.47 & 0.54 & 0.53 & 0.65 & 0.71 & 0.81 \\
         Sparse ORM PPO & 0.46 & 0.51 & 0.51 & 0.55 & 0.59 & 0.67 & 0.79 & 0.83 \\
         Dense ORM PPO & 0.46 & 0.51 & 0.52 & 0.55 & 0.59 & 0.67 & 0.76 & 0.83 \\
         \bottomrule
         Llema$^*$ & 0.40 & 0.62 & 0.54 & 0.69 & \multicolumn{2}{c}{N/A} & \multicolumn{2}{c}{N/A} \\
         RFT  & 0.47 & 0.54 & 0.58 & 0.65 & \multicolumn{2}{c}{N/A} & \multicolumn{2}{c}{N/A} \\
         WizardMath & 0.55 & 0.64 & \multicolumn{2}{c}{N/A} & \multicolumn{2}{c}{N/A} & \multicolumn{2}{c}{N/A} \\
         GPT-3$^{**}$ & 0.2 & 0.31 & \multicolumn{2}{c}{N/A} & 0.39 & 0.55 & 0.71 & NA \\
         GPT-4$^{***}$ & \multicolumn{2}{c}{0.91} & \multicolumn{2}{c}{N/A} & \multicolumn{2}{c}{N/A} & \multicolumn{2}{c}{N/A} \\
        \bottomrule
    \end{tabular}
    \caption{Results when initializing from SFT. EI$_n$ denotes n rounds of expert iteration until convergence with $n=2$ for 7B and $n=2$ for 13B. $^\dagger$Note all reranking is done using an ORM trained with samples from EI$_n$. Results from other works are included on the bottom for reference. N/A stands for not available. $^*$Llema results reported for 7B/34B sizes without fine-tuning. $^{**}$GPT-3 results reported for 7B/175B sizes. $^{***}$GPT-4 size unknown.}
    \label{tab:sft_ppo_results}
\end{table}


\subsection{Results with SFT Initialization}

\begin{figure}[ht]
  \begin{minipage}[t]{0.48\textwidth} 
    \centering
    \includegraphics[scale=0.5]{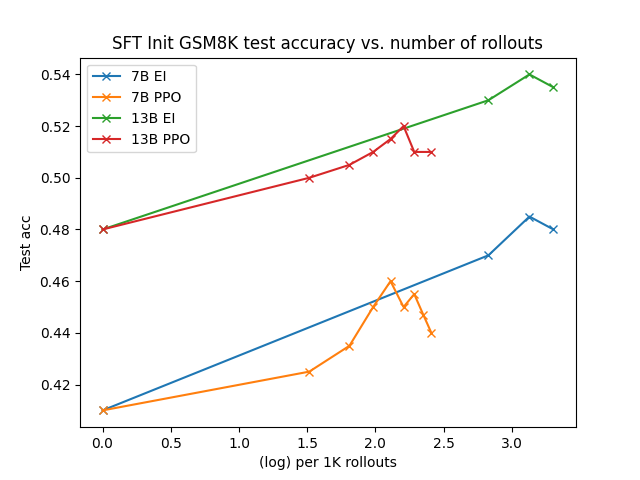}
    \caption{Sample complexities of SFT initialized models on GSM8K. EI achieves better performance than PPO with the same order of magnitude of samples.}
    \label{fig:sft_init_gsm8k_sample_complexity}
  \end{minipage}
  \hfill 
  \begin{minipage}[t]{0.48\textwidth} 
    \centering
    \includegraphics[scale=0.5]{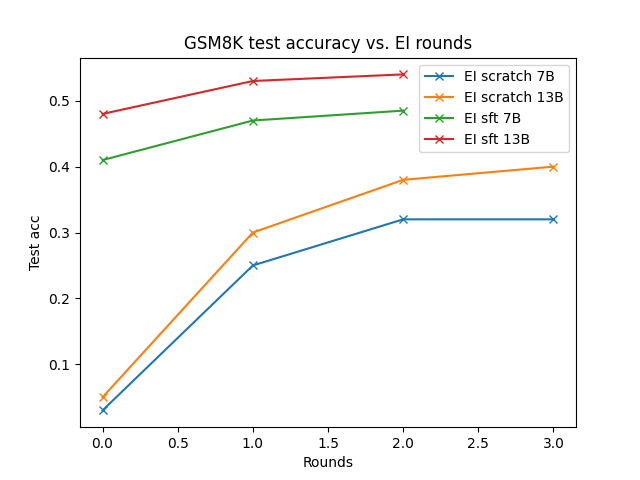}
    \caption{Accuracy of EI models on GSM8K test vs. number of iterations. Performance seems plateaus for SFT initialized models after two iterations. The pretrained checkpoints converge after four iterations.}
    \label{fig:gsm8k_ei_round_acc}
  \end{minipage}
\end{figure}

When given access to SFT data, we first supervise fine-tune Llama-2 models for 4 epochs with a global batch size of 128 and an initial lr of 2e-5 decayed to 2e-7 with a cosine warmup schedule. We call the resulting models \textbf{SFT}. When fine-tuning with PPO we initialize using this checkpoint. In contrast, for both EI and RCRL we generate data with the SFT checkpoint but reset training to start from the pretrained base model. Similarly to \citet{zelikman2022star}, we find this model resetting is crucial for achieving best performance. Results for both 7B and 13B models are reported in Table \ref{tab:sft_ppo_results}.


\textbf{Expert iteration achieves the best performance with competitive sample complexity}

Surprisingly, we find EI achieves the best performance with a maj@1 accuracy of 0.485 and 0.53 on 7B and 13B models respectively. For both model sizes the best greedy accuracy is achieved after $n = 2$ expert iterations (see Fig. \ref{fig:gsm8k_ei_round_acc}), after which performance plateaus. In total, EI gives a sizable improvement of around 7\% over the SFT baseline. Similar gains can be seen in maj@96, rerank@96, and pass@96 scores with.

PPO models underperform EI, with ORM guided PPO giving the biggest improvement of around 5\% over the SFT baseline. Again, maj@96, rerank@96, and pass@96 accuracies show similar improvements. Interestingly, despite further training on top of the SFT initialization, PPO models retain competitive rerank@96 and pass@96 scores when compared to regression we see after further supervised fine-tuning. We believe this is due to the relatively more diverse nature of the exploration dataset used to update the model.

Finally, RCRL models under-perform EI models despite training on EI generated data with an even balance between `[GOOD]' and `[BAD]' step labels. This matches similar results from \cite{Du2023AStudy} which use only sparse labels for the entire rollout. Further, when sampling the RCRL model unconditionally the model often generates the perfectly valid steps following a `[BAD]' label resulting in a correct final answer. These results suggest RCRL models are not correctly learning what constitutes a `[GOOD]' versus `[BAD]'. This suggests RCRL models are unable to usefully incorporate information from partially correct solutions at train time. An ablation (See sec. \ref{sec:rcrl_label_balance} of the appendix) on the ratio of positive to negative labels finds a balanced ratio yields the worst performance, with increasing the amount of positive data leading to better results.


In Figure \ref{fig:sft_init_gsm8k_sample_complexity} we plot the number of model rollouts against model performance in log-scale. PPO models achieve their best accuracies after around 60,000 rollouts while EI models train with an order of magnitude more. However, the resulting train time in both cases is about a day. This is largely due to memory requirements from PPO, resulting in lower rollout throughput and smaller mini-batch sizes at train time. Additionally, in the SFT case we did not experiment with reducing the number of samples from $K = 96$ per question for EI. However, we expect this number can be significantly reduced without impacting performance. For a more thorough investigation of sample complexity requirements, see Figure \ref{fig:no_init_gsm8k_sample_complexity}.

\begin{figure}[ht]
  \begin{minipage}[t]{0.48\textwidth} 
    \centering
    \includegraphics[scale=0.5]{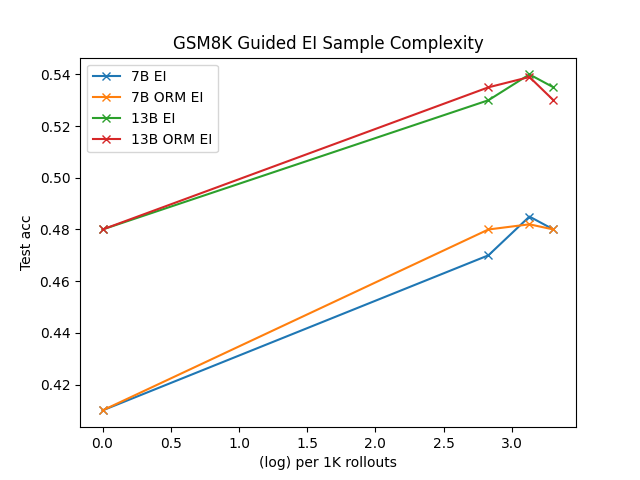}
    \caption{maj@1 scores of EI and ORM aided EI models over the course of training. The ORM improves sample efficiency but not performance.}
    \label{fig:sft-gsm8k-guided-ei-sample-complexity}
  \end{minipage}
  \hfill 
  \begin{minipage}[t]{0.48\textwidth} 
     \centering
    \includegraphics[scale=0.5]{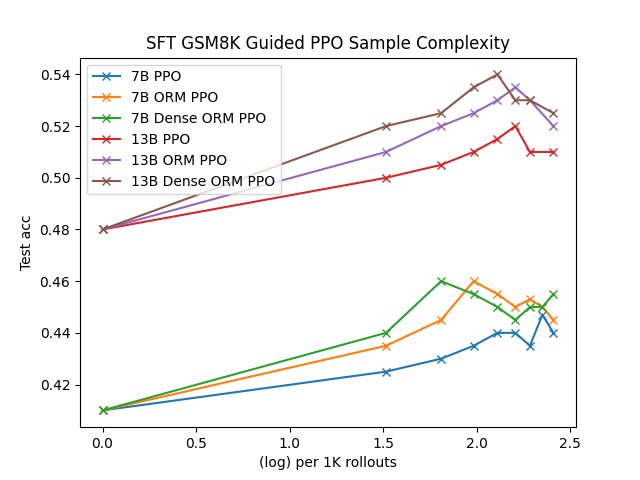}
    \caption{maj@1 scores of PPO and ORM guided PPO models over the course of training. As with EI models, the ORM improves sample efficiency but not performance.}
    \label{fig:gsm8k-ppo-guided-acc}
  \end{minipage}
\end{figure}

\textbf{Extra guidance from ORMs or dense rewards provides little benefit} Overall, the ORM slightly improves PPO performance and negligibly impacts EI performance. For both algorithms it provides an improvement in terms of sample complexity. However, this does not change final performance. See Figures \ref{fig:sft-gsm8k-guided-ei-sample-complexity} and \ref{fig:gsm8k-ppo-guided-acc} which plot the performance against number of model rollouts for differnt reward regimes.

Giving dense rewards at best provides no extra benefit to performance when given either heuristically or via the ORM. Giving a heuristic dense reward even slightly harms model performance relative to the sparse setting. Recall we give intermediate reward by comparing intermediate model generated steps to the reference solution. This likely encourages more overfit to exact solutions in the train set, limiting solution diversity.

\textbf{RL improves maj@1 accuracy without impacting pass@96 performance} Looking at the pass@96 accuracies more closely, we see most similarly sized models are within 3\% of the best result. This demonstrates with enough sampling, most models are able to solve a very similar range of problems. Further, while the pass@96 accuracy of our best EI model initially seems much higher than the SFT checkpoint, this is only because the SFT checkpoint has undergone much more training on a less diverse dataset. Simply supervised fine-tuning for half as many steps results in a checkpoint with maj@1 = 0.36 but pass@96 = 0.76. This further suggests RL training mostly impacts maj@1 accuracy without significantly improving on a pass@n accuracy which can be achieved with a light amount of supervised fine-tuning.

The proximity of pass@96 accuracies among most models is in sharp contrast to the rerank@96 performance. Here we find $EI$ models enjoy around a 5\% lead over other models. At first glance this seems contradictory with relatively similar pass@96 performance. However, we believe a non-trivial percentage of this gap is due to \textbf{overfit of the ORM to the EI model which was used to generate its training data}. 


\subsection{Results with no SFT Initialization}

\begin{table}[t]
    \centering
    \begin{tabular}{@{}lrrrrrrrr@{}}
        \toprule
        & \multicolumn{2}{c}{maj@1} & \multicolumn{2}{c}{maj@n} & \multicolumn{2}{c}{rerank@n$^\dagger$} & \multicolumn{2}{c}{pass@n} \\
        \cmidrule(lr){2-3} \cmidrule(lr){4-5} \cmidrule(lr){6-7} \cmidrule(lr){8-9}
        & 7B & 13B & 7B & 13B & 7B & 13B & 7B & 13B \\
        \midrule
        Prompted & 0.05 & 0.03 & 0.14 & 0.18 & 0.17 & 0.24 & 0.22 & 0.27 \\
         EI$_n$ & 0.31 & 0.4 & 0.35 & 0.47 & 0.39 & 0.63 & 0.45 & \textbf{0.83} \\ 
         ORM EI & 0.28 & 0.37 & 0.33 & 0.43 & 0.37 & 0.59 & 0.42 & 0.76 \\
         Sparse PPO & \textbf{0.32} & \textbf{0.41} & \textbf{0.37} & \textbf{0.48} & \textbf{0.41} & \textbf{0.65} & \textbf{0.5} & \textbf{0.83} \\
         Sparse ORM PPO & 0.29 & 0.38 & 0.34 & 0.44 & 0.4 & 0.62 & 0.49 & 0.81 \\
         Dense ORM PPO & 0.29 & 0.39 & 0.35 & 0.45 & 0.41 & 0.64 & 0.5 & 0.82 \\
        \bottomrule
    \end{tabular}
    \caption{Results for 7B/13B models when \textbf{not} using SFT initialization on GSM8K. Sparse PPO performs slightly better than EIin this setting. $^*$Note all reranking is done using an ORM trained with samples from EI$_n$ model.}
    \label{tab:gsm8k_prompted_ppo_results}
\end{table}

\begin{table}[t]
    \centering
    \begin{tabular}{@{}lrrrrrrrr@{}}
        \toprule
        & \multicolumn{2}{c}{maj@1} & \multicolumn{2}{c}{maj@n} & \multicolumn{2}{c}{rerank@n$^\dagger$} & \multicolumn{2}{c}{pass@n} \\
        \cmidrule(lr){2-3} \cmidrule(lr){4-5} \cmidrule(lr){6-7} \cmidrule(lr){8-9}
        & 7B & 13B & 7B & 13B & 7B & 13B & 7B & 13B \\
        \midrule
        Prompted & 0.06 & 0.05 & 0.2 & 0.25 & 0.24 & 0.29 & 0.3 & 0.36 \\
         EI$_n$ & \textbf{0.58} & \textbf{0.69} & \textbf{0.6} & \textbf{0.75} & \textbf{0.62} & \textbf{0.78} & \textbf{0.70} & \textbf{0.93} \\
         Sparse PPO & 0.44 & 0.51 & 0.55 & 0.66 & 0.58 & 0.73 & 0.72 & 0.89 \\
         Sparse ORM PPO & 0.43 & 0.51 & 0.52 & 0.64 & 0.54 & 0.71 & 0.65 & 0.85 \\
         Dense ORM PPO & 0.44 & 0.52 & 0.51 & 0.63 & 0.55 & 0.73 & 0.67 & 0.85 \\
        \bottomrule
    \end{tabular}
    \caption{Results for 7B/13B models when \textbf{not} using SFT initialization on SVAMP. EI$_n$ denotes the best EI model after $n$ iterations. EI outperforms PPO.}
    \label{tab:svamp_prompted_ppo_results}
\end{table}

\begin{figure}[ht]
  \begin{minipage}[t]{0.48\textwidth} 
    \centering
    \includegraphics[scale=0.5]{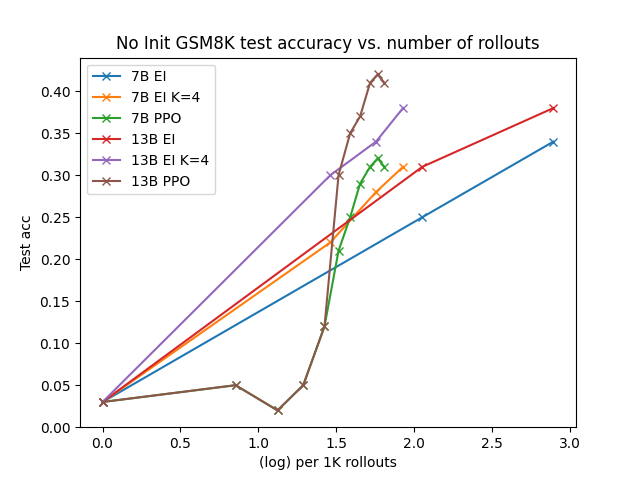}
    \caption{Sample complexities on GSM8K from pretrained initialization.}
    \label{fig:no_init_gsm8k_sample_complexity}
  \end{minipage}
  \hfill 
  \begin{minipage}[t]{0.48\textwidth} 
    \centering
    \includegraphics[scale=0.5]{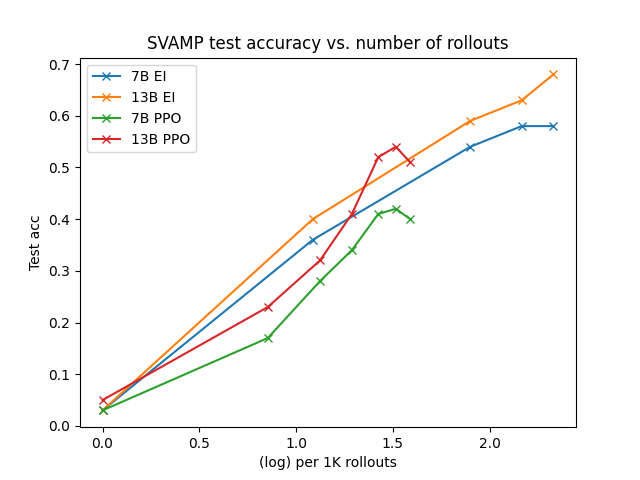}
    \caption{Sample complexities on SVAMP. Surprisingly, EI appears nearly as sample efficient as PPO.}
    \label{fig:svamp_sample_complexity}
  \end{minipage}
\end{figure}

We now consider the case when no SFT data is available for training. For questions in both SVAMP and GSM8K we give pretrained models access to a two-shot prompt with samples drawn from the GSM8K validation set. For EI models, we remove these prompts after the first round of exploration, instead relying on the generated SFT data. As in the case with SFT data, we run both algorithms until performance saturates. For PPO this happens after 250 steps on SVAMP and 1000 steps on GSM8K. For EI, this happens after $n=5$ rounds of exploration and distillation. Results on both datasets are reported in Tables \ref{tab:gsm8k_prompted_ppo_results} and \ref{tab:svamp_prompted_ppo_results}.

\textbf{EI achieves the best performance overall} Even without SFT data, EI achieves the best performance on SVAMP, improving 7B/13B pretrained greedy model accuracies over 50\% from 0.06/0.05 to 0.58/0.69\%, respectively. PPO performs slightly better than EI on GSM8K, improving from 0.05/0.03 to 0.31/0.4. Both algorithms achieve comparable pass@96 scores across modes sizes, further supporting our observations from the SFT regime that EI mostly improves maj@1 scores relative to PPO. The prompted 13B model on GSM8K even attains 0.83 pass@96 accuracy which is close to the 0.84 pass@96 score achieved by the SFT model, despite having no access to SFT data itself.

\textbf{EI has the same sample complexity as PPO} As before we plot the reward versus number of model rollouts for PPO and EI in Figures \ref{fig:no_init_gsm8k_sample_complexity} and \ref{fig:svamp_sample_complexity}. On GSM8K PPO models attain their best maj@1 accuracies after only 30,000 rollouts and on SVAMP even less. Surprisingly, EI models have the same sample complexity as PPO on SVAMP, requiring more samples to converge but also converging to a much higher accuracy. EI still appears to have higher sample complexity on GSM8K, however as noted before this may be due to oversampling each prompt during the exploration phase. To test this, we reduce the number of samples per prompt each round of EI from $K = 96$ to $K = 4$. The resulting EI models require more iterations to converge but require far less total samples, also converging in accuracy only a few percentage points lower than $K = 96$ samples per prompt. With $K = 4$ rollouts per prompt \textbf{EI has the same sample complexity as PPO} on GSM8K.

This is a particularly surprising finding when compared to the performance of EI and PPO on more classical RL problems training a neural network from scratch. Often PPO enjoys far better sample complexity in these settings. One major difference here is the initialization of our student from a pretrained model which imparts a very strong bias on the kind of behaviors and exploration encountered during RL training. Both the extremely small sample complexity and the comparability of EI and PPO in this setting provide more evidence that models are not truly engaging in complex exploration, but instead primarily drawing on what they already know from the pre-training phase.

\subsection{Implementation Details}

It is well known RL training can be quite sensitive to architectural and hyperparameter choices. This is even more so the case for LLM fine-tuning. In this section we ablate and discuss the factors we found most important in our tasks.

\begin{figure}[ht]
  \begin{minipage}[t]{0.48\textwidth} 
    \centering
    \includegraphics[scale=0.5]{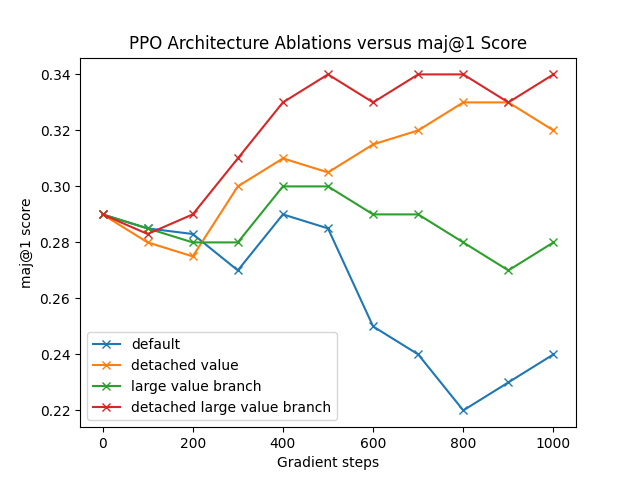}
    \caption{maj@1 performance of PPO fine-tuned models against architectural changes. Note, we initialize training from a 7B SFT model with maj@1 = 0.29.}
    \label{fig:ppo_architecture_ablations}
  \end{minipage}
  \hfill 
  \begin{minipage}[t]{0.48\textwidth} 
    \centering
    \includegraphics[scale=0.5]{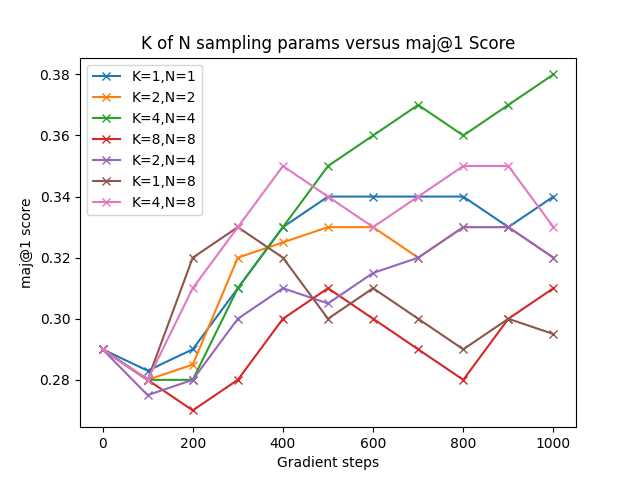}
    \caption{Best K of N sampling parameters versus maj@1 score during training. K=4, N=4 yields a fast runtime and best performance.}
    \label{fig:k_of_n_ablations}
  \end{minipage}
\end{figure}

\textbf{PPO model architecture and training parameters} To save memory we use a joint architecture for the PPO policy and value heads. We found it important to use a relatively large value branch (L=4 transformer layers) and detach the gradients coming from the value branch to the policy trunk. Without detachment we found value gradients interfere with policy gradients, as similarly observed in \citet{Stiennon2020LearningTS}, causing instability with a big update to either branch. See Figure \ref{fig:ppo_architecture_ablations} which compares maj@1 score of a student with a large value branch and detached value gradients versus the default.

Low rank adaptation (LoRA) \citep{Hu2021LoRALA} with rank $r=128$ helped significantly to further stabilize a full layer fine-tuning while still maintaining performance \citep{Sun2023ExploringTI}. A large enough batch size (BS = 256) and a small lr = 1e-6 also helped with stabilization. We additionally experimented with a partial fine-tune of only the top M layers. This saved memory but at the cost of a few percentage points of performance.

We also found a non-trivial KL penalty of $0.05$ to be critical for preventing model collapse after more than a hundred gradient updates. This is in contrast to \citet{Bai2022ConstitutionalAH} who do not see a significant need for the KL constraint. We attribute its importance here to the somewhat unnatural distribution of text found in the the reasoning tasks which consist of broken natural language and computations enclosed in $\texttt{<<x+y=z>>}$ tags. For tasks with distributions closer to pure natural language dialogue, such as those considered in \citet{Bai2022ConstitutionalAH}, the KL constraint seems less necessary. 


\textbf{Sampling parameters affect exploration} We found the best temperature to use for good exploration during PPO training heavily depends on the initialization. When starting from an SFT checkpoint we choose T = 0.7. However, sampling on a high temperature when starting from the pretrained prompted model often results in collapse. In these cases we choose a low temperature (T = 0.2). Potentially better results for PPO could likely be achieved by annealing the exploration temperature over the course of training. We similarly experimented with the sampling temperature used during exploration in EI, ultimately deciding on $T = 1.0$ to maximize solution diversity without sampling too many degenerate solutions.

We also experimented with best K of N (KoN) sampling during PPO training to promote more solution diversity. In this setup the K highest reward samples of N rollouts from a single prompt are kept for training and the rest are discarded. Choosing parameters K $\ll$ N prioritize high reward samples and discard low reward ones, resulting in a training distribution more similar to the curated EI dataset.

However, one important consideration is the impact of the K/N ratio on training time and sample complexity, with smaller ratios taking proportionally longer. For example, K=1,N=8 takes 8 times as long as the default K=1,N=1. Further, we ultimately found little benefit to small K/N ratios with most configurations yielding decreased performance over K=1,N=1. In practice we found setting K=4, N=4 worked best. See Figure \ref{fig:k_of_n_ablations} which compares the performance of various choices of K and N.

\textbf{Model size and initialization affect exploration} We found both the quality of the student initialization and the size of the student significantly affected the type of exploration engaged in during training. In particular \textbf{larger models engaged in more diverse exploration} while \textbf{models with worse generalization engaged in less diverse exploration} (See Appendix Section \ref{sec:diversity}). This in turn directly impacts model performance when trained on exploration data, with models engaging in more diverse exploration improving more from RL training.

\begin{table}[t]
    \centering
    \begin{tabular}{@{}lrrrr@{}}
         & maj@1 & maj@96 & Rerank@96 & pass@96 \\
        \toprule
         SFT$^{2}$ & 0.36 & 0.45 & 0.53 & 0.76 \\
         SFT$^4$ & 0.41 & 0.47 & 0.54 & 0.72 \\
         PPO$^2$ & 0.43 & 0.48 & 0.59 & 0.8 \\
         PPO$^{4}$ & 0.44 & 0.49 & 0.58 & 0.77 \\
         \bottomrule
    \end{tabular}
    \caption{Results for full supervised fine-tune (SFT$^4$), half supervised fine-tune (SFT$^{2}$) and their PPO fine-tunes. Fine-tuning for only two epochs gets pass@96 = 0.76. This decreases to 0.72 with two additional epochs of fine-tuning.}
    \label{tab:weak_sft}
\end{table}

To further examine the observations about overfitting, we supervise fine-tune a Llama-2-7B model for half as many steps than the SFT model reported in Table \ref{tab:sft_ppo_results}. We call the model trained for four epochs SFT$^4$ and the model trained for two epochs SFT$^{2}$. Despite half the training, SFT$^{2}$ has similar Rerank@96 and superior pass@96 scores to SFT$^4$ with the main difference being the maj@1 accuracies. When sampled K = 96 times on each train prompt, SFT$^{2}$ produces on average 3.7 unique correct solutions compared to SFT$^4$ which produces 2.9 unique correct solutions. We also find SFT$^{2}$ benefits significantly more from RL fine-tuning than SFT$^{4}$, jumping from maj@1=0.36 to maj@1=0.43. It's important to note some of this improvement also happens with continued SFT training, however at the cost to model output diversity and pass@96 performance. 

We believe \textbf{RL fine-tuning is less prone to overfitting} when compared to static SFT fine-tuning precisely because of the exploration process which generates its own training data. This results in in more diverse solution paths than the SFT training set, ameliorating overfit. This is also in line with recent work that found RLHF to result in better (out-of-distribution) generalization than SFT on summarization and instruction following tasks~\citep{kirk2023understanding}. This benefit can be found both PPO and EI which have almost 10\% pass@96 improvement over continued SFT (yet a much smaller pass@96 improvement over a light SFT). To support this hypothesis we plot the solution accuracies and diversities of EI models over each iteration in Figures \ref{fig:svamp_ei_round_acc} and \ref{fig:svamp_ei_diversity}, respectively. Figure \ref{fig:svamp_ei_diversity}
also shows larger models generate more diverse solutions.


\section{Discussion and Conclusions}
\label{sec:conclusion}

Our study resulted in the following findings:

\begin{enumerate}
    \item All the tested RL algorithms perform similarly on reasoning tasks, with Expert Iteration performing best in most cases.
    \item Both EI and PPO converge relatively quickly even without supervised fine-tuning, requiring only $\sim$60,000 model rollouts.
    \item Neither algorithm benefits significantly from ORM guidance or a denser reward.
    \item EI and PPO fine-tuning simultaneously improves maj@1 score and pass@n score in contrast with SFT.
\end{enumerate}

The improvement of both maj@1 and pass@n performance noted above is due to the ability of online RL algorithms to dynamically grow diverse sets of training examples via synthetic data generation. This allows for longer training/more gradient updates on the same model without adversely impacting output diversity and pass@n scores. In contrast, SFT training occurs on a static dataset. This limits how much training can occur before maj@1 overfit occurs and output diversity suffers. However, RL training does not significantly improve pass@n score beyond what can be achieved with light supervised fine-tuning. This suggests even with RL training our best models are not discovering solutions beyond what can be discovered with (light) supervised fine-tuning given the same rollout budget.

This observation, taken together with the fast convergence of both online algorithms and the low-impact of ORM guidance and dense rewards, suggests models are not engaging in a significant amount of exploration beyond pretraining/SFT data. Regardless of the type of algorithm used or the quality of the reward, all student models engage in similar exploration, resulting in similar performance.

Crucial in our setting is the usage of a pretrained model imparting a strong exploration prior. Without such a prior, exploration in a high-dimensional textual action space would be impossible. However, this prior also appears to constrain the exploration engaged in at the beginning of training, with additional SFT training only making things worse. We view the discovery of new techniques encouraging complex, rich exploration of reasoning problems as fundamental to progress in LLM reasoning capability. More sophisticted prompting strategies such as Tree of Thought \citep{Yao2023TreeOT} and combining LLM generative abilities with evolutionary algorithms \citep{Lehman2022EvolutionTL} have already begun to make progress in this direction.


In addition to the limited exploration noted above, we also note reasoning environments are entirely deterministic. This is a setting in which EI and RCRL algorithms are already known to work well theoretically \citep{Brandfonbrener2022WhenDR}. PPO enjoys more advantage in environemnts with a high degree of stochasticity. 
We also note prior work in RLHF finds PPO outperforms EI type approaches in human preference satisfaction and instruction following \citep{Gulcehre2023ReinforcedS, Dubois2023AlpacaFarmAS, kirk2023understanding}. Importantly, in our setting we always have a reliable ground truth reward to optimize. However, in RLHF, models must optimize against an unreliable reward model, often resulting in over-optimization \citep{Gao2022ScalingLF}. The relatively superior performance of PPO over EI on RLHF tasks versus reasoning tasks suggests PPO better mitigates such over-optimization. This is not too surprising since PPO training penalizes student models diverging from the initial policy via both its clipped objective and additional KL-constraint. In contrast, EI training has no such protection built in.



\bibliographystyle{plainnat}
\bibliography{rl_review}

\begin{thebibliography}{77}
\providecommand{\natexlab}[1]{#1}
\providecommand{\url}[1]{\texttt{#1}}
\expandafter\ifx\csname urlstyle\endcsname\relax
  \providecommand{\doi}[1]{doi: #1}\else
  \providecommand{\doi}{doi: \begingroup \urlstyle{rm}\Url}\fi

\bibitem[Ammanabrolu and Riedl(2019)]{ammanabrolu-riedl-2019-playing}
Prithviraj Ammanabrolu and Mark Riedl.
\newblock Playing text-adventure games with graph-based deep reinforcement learning.
\newblock In Jill Burstein, Christy Doran, and Thamar Solorio, editors, \emph{Proceedings of the 2019 Conference of the North {A}merican Chapter of the Association for Computational Linguistics: Human Language Technologies, Volume 1 (Long and Short Papers)}, pages 3557--3565, Minneapolis, Minnesota, June 2019. Association for Computational Linguistics.
\newblock \doi{10.18653/v1/N19-1358}.
\newblock URL \url{https://aclanthology.org/N19-1358}.

\bibitem[Anthony et~al.(2017)Anthony, Tian, and Barber]{Anthony2017ThinkingFA}
Thomas~W. Anthony, Zheng Tian, and David Barber.
\newblock Thinking fast and slow with deep learning and tree search.
\newblock In \emph{Neural Information Processing Systems}, 2017.
\newblock URL \url{https://api.semanticscholar.org/CorpusID:19449905}.

\bibitem[Austin et~al.(2021)Austin, Odena, Nye, Bosma, Michalewski, Dohan, Jiang, Cai, Terry, Le, and Sutton]{Austin2021ProgramSW}
Jacob Austin, Augustus Odena, Maxwell Nye, Maarten Bosma, Henryk Michalewski, David Dohan, Ellen Jiang, Carrie~J. Cai, Michael Terry, Quoc~V. Le, and Charles Sutton.
\newblock Program synthesis with large language models.
\newblock \emph{ArXiv}, abs/2108.07732, 2021.
\newblock URL \url{https://api.semanticscholar.org/CorpusID:237142385}.

\bibitem[Azerbayev et~al.(2023)Azerbayev, Schoelkopf, Paster, Santos, McAleer, Jiang, Deng, Biderman, and Welleck]{Azerbayev2023LlemmaAO}
Zhangir Azerbayev, Hailey Schoelkopf, Keiran Paster, Marco~Dos Santos, Stephen McAleer, Albert~Q. Jiang, Jia Deng, Stella Biderman, and Sean Welleck.
\newblock Llemma: An open language model for mathematics.
\newblock \emph{ArXiv}, abs/2310.10631, 2023.
\newblock URL \url{https://api.semanticscholar.org/CorpusID:264172303}.

\bibitem[Bai et~al.(2022)Bai, Kadavath, Kundu, Askell, Kernion, Jones, Chen, Goldie, Mirhoseini, McKinnon, Chen, Olsson, Olah, Hernandez, Drain, Ganguli, Li, Tran-Johnson, Perez, Kerr, Mueller, Ladish, Landau, Ndousse, Lukovsiūtė, Lovitt, Sellitto, Elhage, Schiefer, Mercado, DasSarma, Lasenby, Larson, Ringer, Johnston, Kravec, Showk, Fort, Lanham, Telleen-Lawton, Conerly, Henighan, Hume, Bowman, Hatfield-Dodds, Mann, Amodei, Joseph, McCandlish, Brown, and Kaplan]{Bai2022ConstitutionalAH}
Yuntao Bai, Saurav Kadavath, Sandipan Kundu, Amanda Askell, John Kernion, Andy Jones, Anna Chen, Anna Goldie, Azalia Mirhoseini, Cameron McKinnon, Carol Chen, Catherine Olsson, Christopher Olah, Danny Hernandez, Dawn Drain, Deep Ganguli, Dustin Li, Eli Tran-Johnson, E~Perez, Jamie Kerr, Jared Mueller, Jeff Ladish, J~Landau, Kamal Ndousse, Kamile Lukovsiūtė, Liane Lovitt, Michael Sellitto, Nelson Elhage, Nicholas Schiefer, Noem'i Mercado, Nova DasSarma, Robert Lasenby, Robin Larson, Sam Ringer, Scott Johnston, Shauna Kravec, Sheer~El Showk, Stanislav Fort, Tamera Lanham, Timothy Telleen-Lawton, Tom Conerly, T.~J. Henighan, Tristan Hume, Sam Bowman, Zac Hatfield-Dodds, Benjamin Mann, Dario Amodei, Nicholas Joseph, Sam McCandlish, Tom~B. Brown, and Jared Kaplan.
\newblock Constitutional ai: Harmlessness from ai feedback.
\newblock \emph{ArXiv}, abs/2212.08073, 2022.
\newblock URL \url{https://api.semanticscholar.org/CorpusID:254823489}.

\bibitem[Bakhtin et~al.(2022)Bakhtin, Brown, Dinan, Farina, Flaherty, Fried, Goff, Gray, Hu, Jacob, Komeili, Konath, Kwon, Lerer, Lewis, Miller, Mitts, Renduchintala, Roller, Rowe, Shi, Spisak, Wei, Wu, Zhang, and Zijlstra]{cicero}
Anton Bakhtin, Noam Brown, Emily Dinan, Gabriele Farina, Colin Flaherty, Daniel Fried, Andrew Goff, Jonathan Gray, Hengyuan Hu, Athul~Paul Jacob, Mojtaba Komeili, Karthik Konath, Minae Kwon, Adam Lerer, Mike Lewis, Alexander~H. Miller, Sasha Mitts, Adithya Renduchintala, Stephen Roller, Dirk Rowe, Weiyan Shi, Joe Spisak, Alexander Wei, David Wu, Hugh Zhang, and Markus Zijlstra.
\newblock Human-level play in the game of <i>diplomacy</i> by combining language models with strategic reasoning.
\newblock \emph{Science}, 378\penalty0 (6624):\penalty0 1067--1074, 2022.
\newblock \doi{10.1126/science.ade9097}.
\newblock URL \url{https://www.science.org/doi/abs/10.1126/science.ade9097}.

\bibitem[Berner et~al.(2019)Berner, Brockman, Chan, Cheung, Debiak, Dennison, Farhi, Fischer, Hashme, Hesse, J{\'o}zefowicz, Gray, Olsson, Pachocki, Petrov, de~Oliveira~Pinto, Raiman, Salimans, Schlatter, Schneider, Sidor, Sutskever, Tang, Wolski, and Zhang]{Berner2019Dota2W}
Christopher Berner, Greg Brockman, Brooke Chan, Vicki Cheung, Przemyslaw Debiak, Christy Dennison, David Farhi, Quirin Fischer, Shariq Hashme, Christopher Hesse, Rafal J{\'o}zefowicz, Scott Gray, Catherine Olsson, Jakub~W. Pachocki, Michael Petrov, Henrique~Pond{\'e} de~Oliveira~Pinto, Jonathan Raiman, Tim Salimans, Jeremy Schlatter, Jonas Schneider, Szymon Sidor, Ilya Sutskever, Jie Tang, Filip Wolski, and Susan Zhang.
\newblock Dota 2 with large scale deep reinforcement learning.
\newblock \emph{ArXiv}, abs/1912.06680, 2019.
\newblock URL \url{https://api.semanticscholar.org/CorpusID:209376771}.

\bibitem[Besta et~al.(2023)Besta, Blach, Kubivcek, Gerstenberger, Gianinazzi, Gajda, Lehmann, Podstawski, Niewiadomski, Nyczyk, and Hoefler]{Besta2023GraphOT}
Maciej Besta, Nils Blach, Alevs Kubivcek, Robert Gerstenberger, Lukas Gianinazzi, Joanna Gajda, Tomasz Lehmann, Michal Podstawski, Hubert Niewiadomski, Piotr Nyczyk, and Torsten Hoefler.
\newblock Graph of thoughts: Solving elaborate problems with large language models.
\newblock \emph{ArXiv}, abs/2308.09687, 2023.
\newblock URL \url{https://api.semanticscholar.org/CorpusID:261030303}.

\bibitem[Bhargava et~al.(2023)Bhargava, Chitnis, Geramifard, Sodhani, and Zhang]{Bhargava2023SequenceMI}
Prajjwal Bhargava, Rohan Chitnis, Alborz Geramifard, Shagun Sodhani, and Amy Zhang.
\newblock Sequence modeling is a robust contender for offline reinforcement learning.
\newblock \emph{ArXiv}, abs/2305.14550, 2023.
\newblock URL \url{https://api.semanticscholar.org/CorpusID:258866105}.

\bibitem[Brandfonbrener et~al.(2022)Brandfonbrener, Bietti, Buckman, Laroche, and Bruna]{Brandfonbrener2022WhenDR}
David Brandfonbrener, Alberto Bietti, Jacob Buckman, Romain Laroche, and Joan Bruna.
\newblock When does return-conditioned supervised learning work for offline reinforcement learning?
\newblock \emph{ArXiv}, abs/2206.01079, 2022.
\newblock URL \url{https://api.semanticscholar.org/CorpusID:249282285}.

\bibitem[Carta et~al.(2023)Carta, Romac, Wolf, Lamprier, Sigaud, and Oudeyer]{Carta2023GroundingLL}
Thomas Carta, Cl{\'e}ment Romac, Thomas Wolf, Sylvain Lamprier, Olivier Sigaud, and Pierre-Yves Oudeyer.
\newblock Grounding large language models in interactive environments with online reinforcement learning.
\newblock \emph{ArXiv}, abs/2302.02662, 2023.
\newblock URL \url{https://api.semanticscholar.org/CorpusID:256615643}.

\bibitem[Chen et~al.(2021)Chen, Lu, Rajeswaran, Lee, Grover, Laskin, Abbeel, Srinivas, and Mordatch]{Chen2021DecisionTR}
Lili Chen, Kevin Lu, Aravind Rajeswaran, Kimin Lee, Aditya Grover, Michael Laskin, P.~Abbeel, A.~Srinivas, and Igor Mordatch.
\newblock Decision transformer: Reinforcement learning via sequence modeling.
\newblock In \emph{Neural Information Processing Systems}, 2021.
\newblock URL \url{https://api.semanticscholar.org/CorpusID:235294299}.

\bibitem[Chen et~al.(2022)Chen, Ma, Wang, and Cohen]{Chen2022ProgramOT}
Wenhu Chen, Xueguang Ma, Xinyi Wang, and William~W. Cohen.
\newblock Program of thoughts prompting: Disentangling computation from reasoning for numerical reasoning tasks.
\newblock \emph{ArXiv}, abs/2211.12588, 2022.

\bibitem[Chollet(2019)]{Chollet2019OnTM}
François Chollet.
\newblock On the measure of intelligence.
\newblock \emph{ArXiv}, abs/1911.01547, 2019.
\newblock URL \url{https://api.semanticscholar.org/CorpusID:207870692}.

\bibitem[Chowdhery et~al.(2022)Chowdhery, Narang, Devlin, Bosma, Mishra, Roberts, Barham, Chung, Sutton, Gehrmann, Schuh, Shi, Tsvyashchenko, Maynez, Rao, Barnes, Tay, Shazeer, Prabhakaran, Reif, Du, Hutchinson, Pope, Bradbury, Austin, Isard, Gur-Ari, Yin, Duke, Levskaya, Ghemawat, Dev, Michalewski, Garc{\'i}a, Misra, Robinson, Fedus, Zhou, Ippolito, Luan, Lim, Zoph, Spiridonov, Sepassi, Dohan, Agrawal, Omernick, Dai, Pillai, Pellat, Lewkowycz, Moreira, Child, Polozov, Lee, Zhou, Wang, Saeta, D{\'i}az, Firat, Catasta, Wei, Meier-Hellstern, Eck, Dean, Petrov, and Fiedel]{Chowdhery2022PaLMSL}
Aakanksha Chowdhery, Sharan Narang, Jacob Devlin, Maarten Bosma, Gaurav Mishra, Adam Roberts, Paul Barham, Hyung~Won Chung, Charles Sutton, Sebastian Gehrmann, Parker Schuh, Kensen Shi, Sasha Tsvyashchenko, Joshua Maynez, Abhishek Rao, Parker Barnes, Yi~Tay, Noam~M. Shazeer, Vinodkumar Prabhakaran, Emily Reif, Nan Du, Benton~C. Hutchinson, Reiner Pope, James Bradbury, Jacob Austin, Michael Isard, Guy Gur-Ari, Pengcheng Yin, Toju Duke, Anselm Levskaya, Sanjay Ghemawat, Sunipa Dev, Henryk Michalewski, Xavier Garc{\'i}a, Vedant Misra, Kevin Robinson, Liam Fedus, Denny Zhou, Daphne Ippolito, David Luan, Hyeontaek Lim, Barret Zoph, Alexander Spiridonov, Ryan Sepassi, David Dohan, Shivani Agrawal, Mark Omernick, Andrew~M. Dai, Thanumalayan~Sankaranarayana Pillai, Marie Pellat, Aitor Lewkowycz, Erica Moreira, Rewon Child, Oleksandr Polozov, Katherine Lee, Zongwei Zhou, Xuezhi Wang, Brennan Saeta, Mark D{\'i}az, Orhan Firat, Michele Catasta, Jason Wei, Kathleen~S. Meier-Hellstern, Douglas Eck, Jeff Dean, Slav Petrov,
  and Noah Fiedel.
\newblock Palm: Scaling language modeling with pathways.
\newblock \emph{J. Mach. Learn. Res.}, 24:\penalty0 240:1--240:113, 2022.
\newblock URL \url{https://api.semanticscholar.org/CorpusID:247951931}.

\bibitem[Christiano et~al.(2017)Christiano, Leike, Brown, Martic, Legg, and Amodei]{christiano2017deep}
Paul~F Christiano, Jan Leike, Tom Brown, Miljan Martic, Shane Legg, and Dario Amodei.
\newblock Deep reinforcement learning from human preferences.
\newblock \emph{Advances in neural information processing systems}, 30, 2017.

\bibitem[Cobbe et~al.(2021)Cobbe, Kosaraju, Bavarian, Chen, Jun, Kaiser, Plappert, Tworek, Hilton, Nakano, Hesse, and Schulman]{Cobbe2021TrainingVT}
Karl Cobbe, Vineet Kosaraju, Mohammad Bavarian, Mark Chen, Heewoo Jun, Lukasz Kaiser, Matthias Plappert, Jerry Tworek, Jacob Hilton, Reiichiro Nakano, Christopher Hesse, and John Schulman.
\newblock Training verifiers to solve math word problems.
\newblock \emph{ArXiv}, abs/2110.14168, 2021.
\newblock URL \url{https://api.semanticscholar.org/CorpusID:239998651}.

\bibitem[Dohan et~al.(2022)Dohan, Xu, Lewkowycz, Austin, Bieber, Lopes, Wu, Michalewski, Saurous, Sohl-Dickstein, Murphy, and Sutton]{Dohan2022LanguageMC}
David Dohan, Winnie Xu, Aitor Lewkowycz, Jacob Austin, David Bieber, Raphael~Gontijo Lopes, Yuhuai Wu, Henryk Michalewski, Rif~A. Saurous, Jascha~Narain Sohl-Dickstein, Kevin Murphy, and Charles Sutton.
\newblock Language model cascades.
\newblock \emph{ArXiv}, abs/2207.10342, 2022.

\bibitem[Dong et~al.(2023)Dong, Xiong, Goyal, Pan, Diao, Zhang, Shum, and Zhang]{Dong2023RAFTRR}
Hanze Dong, Wei Xiong, Deepanshu Goyal, Rui Pan, Shizhe Diao, Jipeng Zhang, Kashun Shum, and T.~Zhang.
\newblock Raft: Reward ranked finetuning for generative foundation model alignment.
\newblock \emph{ArXiv}, abs/2304.06767, 2023.
\newblock URL \url{https://api.semanticscholar.org/CorpusID:258170300}.

\bibitem[Du et~al.(2023)Du, Havrilla, Sukhbaatar, Abbeel, and Raileanu]{Du2023AStudy}
Yuqing Du, Alexander Havrilla, Sainbayar Sukhbaatar, Pieter Abbeel, and Roberta Raileanu.
\newblock A study on improving reasoning in language models.
\newblock In \emph{I Can't Believe It's Not Better Workshop: Failure Modes in the Age of Foundation Models}, 2023.
\newblock URL \url{https://openreview.net/forum?id=tCZFmDyPFm}.

\bibitem[Dubois et~al.(2023)Dubois, Li, Taori, Zhang, Gulrajani, Ba, Guestrin, Liang, and Hashimoto]{Dubois2023AlpacaFarmAS}
Yann Dubois, Xuechen Li, Rohan Taori, Tianyi Zhang, Ishaan Gulrajani, Jimmy Ba, Carlos Guestrin, Percy Liang, and Tatsunori Hashimoto.
\newblock Alpacafarm: A simulation framework for methods that learn from human feedback.
\newblock \emph{ArXiv}, abs/2305.14387, 2023.
\newblock URL \url{https://api.semanticscholar.org/CorpusID:258865545}.

\bibitem[Gao et~al.(2021)Gao, Tow, Biderman, Black, DiPofi, Foster, Golding, Hsu, McDonell, Muennighoff, Phang, Reynolds, Tang, Thite, Wang, Wang, and Zou]{eval-harness}
Leo Gao, Jonathan Tow, Stella Biderman, Sid Black, Anthony DiPofi, Charles Foster, Laurence Golding, Jeffrey Hsu, Kyle McDonell, Niklas Muennighoff, Jason Phang, Laria Reynolds, Eric Tang, Anish Thite, Ben Wang, Kevin Wang, and Andy Zou.
\newblock A framework for few-shot language model evaluation, September 2021.
\newblock URL \url{https://doi.org/10.5281/zenodo.5371628}.

\bibitem[Gao et~al.(2022)Gao, Schulman, and Hilton]{Gao2022ScalingLF}
Leo Gao, John Schulman, and Jacob Hilton.
\newblock Scaling laws for reward model overoptimization.
\newblock In \emph{International Conference on Machine Learning}, 2022.
\newblock URL \url{https://api.semanticscholar.org/CorpusID:252992904}.

\bibitem[Glaese et~al.(2022)Glaese, McAleese, Trkebacz, Aslanides, Firoiu, Ewalds, Rauh, Weidinger, Chadwick, Thacker, Campbell-Gillingham, Uesato, Huang, Comanescu, Yang, See, Dathathri, Greig, Chen, Fritz, Elias, Green, Mokr'a, Fernando, Wu, Foley, Young, Gabriel, Isaac, Mellor, Hassabis, Kavukcuoglu, Hendricks, and Irving]{Glaese2022ImprovingAO}
Amelia Glaese, Nathan McAleese, Maja Trkebacz, John Aslanides, Vlad Firoiu, Timo Ewalds, Maribeth Rauh, Laura Weidinger, Martin Chadwick, Phoebe Thacker, Lucy Campbell-Gillingham, Jonathan Uesato, Po-Sen Huang, Ramona Comanescu, Fan Yang, A.~See, Sumanth Dathathri, Rory Greig, Charlie Chen, Doug Fritz, Jaume~Sanchez Elias, Richard Green, Sovna Mokr'a, Nicholas Fernando, Boxi Wu, Rachel Foley, Susannah Young, Iason Gabriel, William~S. Isaac, John F.~J. Mellor, Demis Hassabis, Koray Kavukcuoglu, Lisa~Anne Hendricks, and Geoffrey Irving.
\newblock Improving alignment of dialogue agents via targeted human judgements.
\newblock \emph{ArXiv}, abs/2209.14375, 2022.
\newblock URL \url{https://api.semanticscholar.org/CorpusID:252596089}.

\bibitem[Gulcehre et~al.(2023)Gulcehre, Paine, Srinivasan, Konyushkova, Weerts, Sharma, Siddhant, Ahern, Wang, Gu, Macherey, Doucet, Firat, and de~Freitas]{Gulcehre2023ReinforcedS}
Caglar Gulcehre, Tom~Le Paine, Srivatsan Srinivasan, Ksenia Konyushkova, Lotte Weerts, Abhishek Sharma, Aditya Siddhant, Alexa Ahern, Miaosen Wang, Chenjie Gu, Wolfgang Macherey, A.~Doucet, Orhan Firat, and Nando de~Freitas.
\newblock Reinforced self-training (rest) for language modeling.
\newblock \emph{ArXiv}, abs/2308.08998, 2023.
\newblock URL \url{https://api.semanticscholar.org/CorpusID:261031028}.

\bibitem[Hendrycks et~al.(2021{\natexlab{a}})Hendrycks, Burns, Kadavath, Arora, Basart, Tang, Song, and Steinhardt]{hendrycks2021measuring}
Dan Hendrycks, Collin Burns, Saurav Kadavath, Akul Arora, Steven Basart, Eric Tang, Dawn Song, and Jacob Steinhardt.
\newblock Measuring mathematical problem solving with the math dataset, 2021{\natexlab{a}}.

\bibitem[Hendrycks et~al.(2021{\natexlab{b}})Hendrycks, Burns, Kadavath, Arora, Basart, Tang, Song, and Steinhardt]{Hendrycks2021MeasuringMP}
Dan Hendrycks, Collin Burns, Saurav Kadavath, Akul Arora, Steven Basart, Eric Tang, Dawn~Xiaodong Song, and Jacob Steinhardt.
\newblock Measuring mathematical problem solving with the math dataset.
\newblock \emph{ArXiv}, abs/2103.03874, 2021{\natexlab{b}}.
\newblock URL \url{https://api.semanticscholar.org/CorpusID:232134851}.

\bibitem[Hu et~al.(2021)Hu, Shen, Wallis, Allen-Zhu, Li, Wang, and Chen]{Hu2021LoRALA}
J.~Edward Hu, Yelong Shen, Phillip Wallis, Zeyuan Allen-Zhu, Yuanzhi Li, Shean Wang, and Weizhu Chen.
\newblock Lora: Low-rank adaptation of large language models.
\newblock \emph{ArXiv}, abs/2106.09685, 2021.
\newblock URL \url{https://api.semanticscholar.org/CorpusID:235458009}.

\bibitem[Huang et~al.(2022)Huang, Gu, Hou, Wu, Wang, Yu, and Han]{Huang2022LargeLM}
Jiaxin Huang, Shixiang~Shane Gu, Le~Hou, Yuexin Wu, Xuezhi Wang, Hongkun Yu, and Jiawei Han.
\newblock Large language models can self-improve.
\newblock \emph{ArXiv}, abs/2210.11610, 2022.

\bibitem[Jiang et~al.(2020)Jiang, Grefenstette, and Rockt{\"a}schel]{Jiang2020PrioritizedLR}
Minqi Jiang, Edward Grefenstette, and Tim Rockt{\"a}schel.
\newblock Prioritized level replay.
\newblock In \emph{International Conference on Machine Learning}, 2020.
\newblock URL \url{https://api.semanticscholar.org/CorpusID:222208809}.

\bibitem[Kirk et~al.(2023)Kirk, Mediratta, Nalmpantis, Luketina, Hambro, Grefenstette, and Raileanu]{kirk2023understanding}
Robert Kirk, Ishita Mediratta, Christoforos Nalmpantis, Jelena Luketina, Eric Hambro, Edward Grefenstette, and Roberta Raileanu.
\newblock Understanding the effects of rlhf on llm generalisation and diversity.
\newblock \emph{arXiv preprint arXiv:2310.06452}, 2023.

\bibitem[Le et~al.(2022)Le, Wang, Gotmare, Savarese, and Hoi]{Le2022CodeRLMC}
Hung Le, Yue Wang, Akhilesh~Deepak Gotmare, Silvio Savarese, and Steven C.~H. Hoi.
\newblock Coderl: Mastering code generation through pretrained models and deep reinforcement learning.
\newblock \emph{ArXiv}, abs/2207.01780, 2022.
\newblock URL \url{https://api.semanticscholar.org/CorpusID:250280117}.

\bibitem[Lehman et~al.(2022)Lehman, Gordon, Jain, Ndousse, Yeh, and Stanley]{Lehman2022EvolutionTL}
Joel Lehman, Jonathan Gordon, Shawn Jain, Kamal Ndousse, Cathy Yeh, and Kenneth~O. Stanley.
\newblock Evolution through large models.
\newblock 2022.
\newblock URL \url{https://api.semanticscholar.org/CorpusID:249848020}.

\bibitem[Lewkowycz et~al.(2022)Lewkowycz, Andreassen, Dohan, Dyer, Michalewski, Ramasesh, Slone, Anil, Schlag, Gutman-Solo, Wu, Neyshabur, Gur-Ari, and Misra]{Lewkowycz2022SolvingQR}
Aitor Lewkowycz, Anders Andreassen, David Dohan, Ethan Dyer, Henryk Michalewski, Vinay~Venkatesh Ramasesh, Ambrose Slone, Cem Anil, Imanol Schlag, Theo Gutman-Solo, Yuhuai Wu, Behnam Neyshabur, Guy Gur-Ari, and Vedant Misra.
\newblock Solving quantitative reasoning problems with language models.
\newblock \emph{ArXiv}, abs/2206.14858, 2022.
\newblock URL \url{https://api.semanticscholar.org/CorpusID:250144408}.

\bibitem[Li et~al.(2023)Li, Allal, Zi, Muennighoff, Kocetkov, Mou, Marone, Akiki, Li, Chim, Liu, Zheltonozhskii, Zhuo, Wang, Dehaene, Davaadorj, Lamy-Poirier, Monteiro, Shliazhko, Gontier, Meade, Zebaze, Yee, Umapathi, Zhu, Lipkin, Oblokulov, Wang, Murthy, Stillerman, Patel, Abulkhanov, Zocca, Dey, Zhang, Fahmy, Bhattacharyya, Yu, Singh, Luccioni, Villegas, Kunakov, Zhdanov, Romero, Lee, Timor, Ding, Schlesinger, Schoelkopf, Ebert, Dao, Mishra, Gu, Robinson, Anderson, Dolan-Gavitt, Contractor, Reddy, Fried, Bahdanau, Jernite, Ferrandis, Hughes, Wolf, Guha, von Werra, and de~Vries]{Li2023StarCoderMT}
Raymond Li, Loubna~Ben Allal, Yangtian Zi, Niklas Muennighoff, Denis Kocetkov, Chenghao Mou, Marc Marone, Christopher Akiki, Jia Li, Jenny Chim, Qian Liu, Evgenii Zheltonozhskii, Terry~Yue Zhuo, Thomas Wang, Olivier Dehaene, Mishig Davaadorj, Joel Lamy-Poirier, Jo{\~a}o Monteiro, Oleh Shliazhko, Nicolas Gontier, Nicholas Meade, Armel Zebaze, Ming-Ho Yee, Logesh~Kumar Umapathi, Jian Zhu, Benjamin Lipkin, Muhtasham Oblokulov, Zhiruo Wang, Rudra Murthy, Jason Stillerman, Siva~Sankalp Patel, Dmitry Abulkhanov, Marco Zocca, Manan Dey, Zhihan Zhang, Nourhan Fahmy, Urvashi Bhattacharyya, W.~Yu, Swayam Singh, Sasha Luccioni, Paulo Villegas, Maxim Kunakov, Fedor Zhdanov, Manuel Romero, Tony Lee, Nadav Timor, Jennifer Ding, Claire Schlesinger, Hailey Schoelkopf, Jana Ebert, Tri Dao, Mayank Mishra, Alexander Gu, Jennifer Robinson, Carolyn~Jane Anderson, Brendan Dolan-Gavitt, Danish Contractor, Siva Reddy, Daniel Fried, Dzmitry Bahdanau, Yacine Jernite, Carlos~Mu{\~n}oz Ferrandis, Sean~M. Hughes, Thomas Wolf, Arjun Guha,
  Leandro von Werra, and Harm de~Vries.
\newblock Starcoder: may the source be with you!
\newblock \emph{ArXiv}, abs/2305.06161, 2023.
\newblock URL \url{https://api.semanticscholar.org/CorpusID:258588247}.

\bibitem[Liang et~al.(2022)Liang, Bommasani, Lee, Tsipras, Soylu, Yasunaga, Zhang, Narayanan, Wu, Kumar, Newman, Yuan, Yan, Zhang, Cosgrove, Manning, R{\'e}, Acosta-Navas, Hudson, Zelikman, Durmus, Ladhak, Rong, Ren, Yao, Wang, Santhanam, Orr, Zheng, Y{\"u}ksekg{\"o}n{\"u}l, Suzgun, Kim, Guha, Chatterji, Khattab, Henderson, Huang, Chi, Xie, Santurkar, Ganguli, Hashimoto, Icard, Zhang, Chaudhary, Wang, Li, Mai, Zhang, and Koreeda]{Liang2022HolisticEO}
Percy Liang, Rishi Bommasani, Tony Lee, Dimitris Tsipras, Dilara Soylu, Michihiro Yasunaga, Yian Zhang, Deepak Narayanan, Yuhuai Wu, Ananya Kumar, Benjamin Newman, Binhang Yuan, Bobby Yan, Ce~Zhang, Christian Cosgrove, Christopher~D. Manning, Christopher R{\'e}, Diana Acosta-Navas, Drew~A. Hudson, E.~Zelikman, Esin Durmus, Faisal Ladhak, Frieda Rong, Hongyu Ren, Huaxiu Yao, Jue Wang, Keshav Santhanam, Laurel~J. Orr, Lucia Zheng, Mert Y{\"u}ksekg{\"o}n{\"u}l, Mirac Suzgun, Nathan Kim, Neel Guha, Niladri~S. Chatterji, O.~Khattab, Peter Henderson, Qian Huang, Ryan Chi, Sang~Michael Xie, Shibani Santurkar, Surya Ganguli, Tatsunori Hashimoto, Thomas Icard, Tianyi Zhang, Vishrav Chaudhary, William Wang, Xuechen Li, Yifan Mai, Yuhui Zhang, and Yuta Koreeda.
\newblock Holistic evaluation of language models.
\newblock \emph{ArXiv}, abs/2211.09110, 2022.
\newblock URL \url{https://api.semanticscholar.org/CorpusID:263423935}.

\bibitem[Lightman et~al.(2023)Lightman, Kosaraju, Burda, Edwards, Baker, Lee, Leike, Schulman, Sutskever, and Cobbe]{Lightman2023LetsVS}
Hunter Lightman, Vineet Kosaraju, Yura Burda, Harrison Edwards, Bowen Baker, Teddy Lee, Jan Leike, John Schulman, Ilya Sutskever, and Karl Cobbe.
\newblock Let's verify step by step.
\newblock \emph{ArXiv}, abs/2305.20050, 2023.
\newblock URL \url{https://api.semanticscholar.org/CorpusID:258987659}.

\bibitem[Liu et~al.(2022)Liu, Hallinan, Lu, He, Welleck, Hajishirzi, and Choi]{Liu2022RainierRK}
Jiacheng Liu, Skyler Hallinan, Ximing Lu, Pengfei He, Sean Welleck, Hannaneh Hajishirzi, and Yejin Choi.
\newblock Rainier: Reinforced knowledge introspector for commonsense question answering.
\newblock \emph{ArXiv}, abs/2210.03078, 2022.
\newblock URL \url{https://api.semanticscholar.org/CorpusID:252735191}.

\bibitem[Lu et~al.(2022)Lu, Welleck, Jiang, Hessel, Qin, West, Ammanabrolu, and Choi]{Lu2022QuarkCT}
Ximing Lu, Sean Welleck, Liwei Jiang, Jack Hessel, Lianhui Qin, Peter West, Prithviraj Ammanabrolu, and Yejin Choi.
\newblock Quark: Controllable text generation with reinforced unlearning.
\newblock \emph{ArXiv}, abs/2205.13636, 2022.
\newblock URL \url{https://api.semanticscholar.org/CorpusID:249152301}.

\bibitem[Luo et~al.(2023)Luo, Sun, Xu, Zhao, Lou, Tao, Geng, Lin, Chen, and Zhang]{Luo2023WizardMathEM}
Haipeng Luo, Qingfeng Sun, Can Xu, Pu~Zhao, Jian-Guang Lou, Chongyang Tao, Xiubo Geng, Qingwei Lin, Shifeng Chen, and Dongmei Zhang.
\newblock Wizardmath: Empowering mathematical reasoning for large language models via reinforced evol-instruct.
\newblock \emph{ArXiv}, abs/2308.09583, 2023.
\newblock URL \url{https://api.semanticscholar.org/CorpusID:261030818}.

\bibitem[Mialon et~al.(2023)Mialon, Fourrier, Swift, Wolf, LeCun, and Scialom]{Mialon2023GAIAAB}
Gr{\'e}goire Mialon, Cl{\'e}mentine Fourrier, Craig Swift, Thomas Wolf, Yann~Andr{\'e} LeCun, and Thomas Scialom.
\newblock Gaia: a benchmark for general ai assistants.
\newblock 2023.
\newblock URL \url{https://api.semanticscholar.org/CorpusID:265351664}.

\bibitem[Mishra et~al.(2021)Mishra, Khashabi, Baral, and Hajishirzi]{Mishra2021CrossTaskGV}
Swaroop Mishra, Daniel Khashabi, Chitta Baral, and Hannaneh Hajishirzi.
\newblock Cross-task generalization via natural language crowdsourcing instructions.
\newblock In \emph{Annual Meeting of the Association for Computational Linguistics}, 2021.
\newblock URL \url{https://api.semanticscholar.org/CorpusID:237421373}.

\bibitem[Mishra et~al.(2022)Mishra, Lu, and Kalyan]{Mishra2022LilaAU}
Swaroop Mishra, Pan Lu, and A.~Kalyan.
\newblock Lila: A unified benchmark for mathematical reasoning.
\newblock 2022.
\newblock URL \url{https://api.semanticscholar.org/CorpusID:257405677}.

\bibitem[OpenAI(2023)]{OpenAI2023GPT4TR}
OpenAI.
\newblock Gpt-4 technical report.
\newblock \emph{ArXiv}, abs/2303.08774, 2023.
\newblock URL \url{https://api.semanticscholar.org/CorpusID:257532815}.

\bibitem[Ouyang et~al.(2022)Ouyang, Wu, Jiang, Almeida, Wainwright, Mishkin, Zhang, Agarwal, Slama, Ray, Schulman, Hilton, Kelton, Miller, Simens, Askell, Welinder, Christiano, Leike, and Lowe]{Ouyang2022TrainingLM}
Long Ouyang, Jeff Wu, Xu~Jiang, Diogo Almeida, Carroll~L. Wainwright, Pamela Mishkin, Chong Zhang, Sandhini Agarwal, Katarina Slama, Alex Ray, John Schulman, Jacob Hilton, Fraser Kelton, Luke~E. Miller, Maddie Simens, Amanda Askell, Peter Welinder, Paul~Francis Christiano, Jan Leike, and Ryan~J. Lowe.
\newblock Training language models to follow instructions with human feedback.
\newblock \emph{ArXiv}, abs/2203.02155, 2022.
\newblock URL \url{https://api.semanticscholar.org/CorpusID:246426909}.

\bibitem[Patel et~al.(2021)Patel, Bhattamishra, and Goyal]{patel2021nlp}
Arkil Patel, Satwik Bhattamishra, and Navin Goyal.
\newblock Are nlp models really able to solve simple math word problems?, 2021.

\bibitem[Peng et~al.(2021)Peng, Li, Wiegreffe, and Riedl]{peng2021inferring}
Xiangyu Peng, Siyan Li, Sarah Wiegreffe, and Mark Riedl.
\newblock Inferring the reader: Guiding automated story generation with commonsense reasoning, 2021.

\bibitem[Qin et~al.(2023)Qin, Liang, Ye, Zhu, Yan, Lu, Lin, Cong, Tang, Qian, Zhao, Tian, Xie, Zhou, Gerstein, Li, Liu, and Sun]{Qin2023ToolLLMFL}
Yujia Qin, Shi Liang, Yining Ye, Kunlun Zhu, Lan Yan, Ya-Ting Lu, Yankai Lin, Xin Cong, Xiangru Tang, Bill Qian, Sihan Zhao, Runchu Tian, Ruobing Xie, Jie Zhou, Marc~H. Gerstein, Dahai Li, Zhiyuan Liu, and Maosong Sun.
\newblock Toolllm: Facilitating large language models to master 16000+ real-world apis.
\newblock \emph{ArXiv}, abs/2307.16789, 2023.
\newblock URL \url{https://api.semanticscholar.org/CorpusID:260334759}.

\bibitem[Rafailov et~al.(2023)Rafailov, Sharma, Mitchell, Ermon, Manning, and Finn]{Rafailov2023DirectPO}
Rafael Rafailov, Archit Sharma, Eric Mitchell, Stefano Ermon, Christopher~D. Manning, and Chelsea Finn.
\newblock Direct preference optimization: Your language model is secretly a reward model.
\newblock \emph{ArXiv}, abs/2305.18290, 2023.
\newblock URL \url{https://api.semanticscholar.org/CorpusID:258959321}.

\bibitem[Ramamurthy et~al.(2022)Ramamurthy, Ammanabrolu, Brantley, Hessel, Sifa, Bauckhage, Hajishirzi, and Choi]{Ramamurthy2022IsRL}
Rajkumar Ramamurthy, Prithviraj Ammanabrolu, Kiant{\'e} Brantley, Jack Hessel, Rafet Sifa, Christian Bauckhage, Hannaneh Hajishirzi, and Yejin Choi.
\newblock Is reinforcement learning (not) for natural language processing?: Benchmarks, baselines, and building blocks for natural language policy optimization.
\newblock \emph{ArXiv}, abs/2210.01241, 2022.
\newblock URL \url{https://api.semanticscholar.org/CorpusID:252693405}.

\bibitem[Rein et~al.(2023)Rein, Hou, Stickland, Petty, Pang, Dirani, Michael, and Bowman]{Rein2023GPQAAG}
David Rein, Betty~Li Hou, Asa~Cooper Stickland, Jackson Petty, Richard~Yuanzhe Pang, Julien Dirani, Julian Michael, and Samuel~R. Bowman.
\newblock Gpqa: A graduate-level google-proof q\&a benchmark.
\newblock \emph{ArXiv}, abs/2311.12022, 2023.
\newblock URL \url{https://api.semanticscholar.org/CorpusID:265295009}.

\bibitem[Rozière et~al.(2023)Rozière, Gehring, Gloeckle, Sootla, Gat, Tan, Adi, Liu, Remez, Rapin, Kozhevnikov, Evtimov, Bitton, Bhatt, Ferrer, Grattafiori, Xiong, Défossez, Copet, Azhar, Touvron, Martin, Usunier, Scialom, and Synnaeve]{rozière2023code}
Baptiste Rozière, Jonas Gehring, Fabian Gloeckle, Sten Sootla, Itai Gat, Xiaoqing~Ellen Tan, Yossi Adi, Jingyu Liu, Tal Remez, Jérémy Rapin, Artyom Kozhevnikov, Ivan Evtimov, Joanna Bitton, Manish Bhatt, Cristian~Canton Ferrer, Aaron Grattafiori, Wenhan Xiong, Alexandre Défossez, Jade Copet, Faisal Azhar, Hugo Touvron, Louis Martin, Nicolas Usunier, Thomas Scialom, and Gabriel Synnaeve.
\newblock Code llama: Open foundation models for code, 2023.

\bibitem[Salimans and Chen(2018)]{Salimans2018LearningMR}
Tim Salimans and Richard~J. Chen.
\newblock Learning montezuma's revenge from a single demonstration.
\newblock \emph{ArXiv}, abs/1812.03381, 2018.
\newblock URL \url{https://api.semanticscholar.org/CorpusID:54463584}.

\bibitem[Sawada et~al.(2023)Sawada, Paleka, Havrilla, Tadepalli, Vidas, Kranias, Nay, Gupta, and Komatsuzaki]{Sawada2023ARBAR}
Tomohiro Sawada, Daniel Paleka, Alex Havrilla, Pranav Tadepalli, Paula Vidas, Alexander Kranias, John~J. Nay, Kshitij Gupta, and Aran Komatsuzaki.
\newblock Arb: Advanced reasoning benchmark for large language models.
\newblock \emph{ArXiv}, abs/2307.13692, 2023.
\newblock URL \url{https://api.semanticscholar.org/CorpusID:260155126}.

\bibitem[Schick et~al.(2023)Schick, Dwivedi-Yu, Dess{\`i}, Raileanu, Lomeli, Zettlemoyer, Cancedda, and Scialom]{Schick2023ToolformerLM}
Timo Schick, Jane Dwivedi-Yu, Roberto Dess{\`i}, Roberta Raileanu, Maria Lomeli, Luke Zettlemoyer, Nicola Cancedda, and Thomas Scialom.
\newblock Toolformer: Language models can teach themselves to use tools.
\newblock \emph{ArXiv}, abs/2302.04761, 2023.
\newblock URL \url{https://api.semanticscholar.org/CorpusID:256697342}.

\bibitem[Schulman et~al.(2017)Schulman, Wolski, Dhariwal, Radford, and Klimov]{Schulman2017ProximalPO}
John Schulman, Filip Wolski, Prafulla Dhariwal, Alec Radford, and Oleg Klimov.
\newblock Proximal policy optimization algorithms.
\newblock \emph{ArXiv}, abs/1707.06347, 2017.
\newblock URL \url{https://api.semanticscholar.org/CorpusID:28695052}.

\bibitem[Sennrich et~al.(2015)Sennrich, Haddow, and Birch]{Sennrich2015ImprovingNM}
Rico Sennrich, Barry Haddow, and Alexandra Birch.
\newblock Improving neural machine translation models with monolingual data.
\newblock \emph{ArXiv}, abs/1511.06709, 2015.
\newblock URL \url{https://api.semanticscholar.org/CorpusID:15600925}.

\bibitem[Shen et~al.(2023)Shen, Zhang, Chen, Zan, Geng, Fu, Zeng, Yu, Ji, Zhao, Guo, and Wang]{Shen2023PanGuCoder2BL}
Bo~Shen, Jiaxin Zhang, Taihong Chen, Daoguang Zan, Bing Geng, An~Fu, Muhan Zeng, Ailun Yu, Jichuan Ji, Jingyang Zhao, Yuenan Guo, and Qianxiang Wang.
\newblock Pangu-coder2: Boosting large language models for code with ranking feedback.
\newblock \emph{ArXiv}, abs/2307.14936, 2023.
\newblock URL \url{https://api.semanticscholar.org/CorpusID:260202985}.

\bibitem[Silver et~al.(2017)Silver, Hubert, Schrittwieser, Antonoglou, Lai, Guez, Lanctot, Sifre, Kumaran, Graepel, Lillicrap, Simonyan, and Hassabis]{Silver2017MasteringCA}
David Silver, Thomas Hubert, Julian Schrittwieser, Ioannis Antonoglou, Matthew Lai, Arthur Guez, Marc Lanctot, L.~Sifre, Dharshan Kumaran, Thore Graepel, Timothy~P. Lillicrap, Karen Simonyan, and Demis Hassabis.
\newblock Mastering chess and shogi by self-play with a general reinforcement learning algorithm.
\newblock \emph{ArXiv}, abs/1712.01815, 2017.
\newblock URL \url{https://api.semanticscholar.org/CorpusID:33081038}.

\bibitem[Srivastava et~al.(2022)Srivastava, Rastogi, Rao, Shoeb, Abid, Fisch, Brown, Santoro, Gupta, Garriga-Alonso, Kluska, Lewkowycz, Agarwal, Power, Ray, Warstadt, Kocurek, Safaya, Tazarv, Xiang, Parrish, Nie, Hussain, Askell, Dsouza, Slone, Rahane, Iyer, Andreassen, Madotto, Santilli, Stuhlmuller, Dai, La, Lampinen, Zou, Jiang, Chen, Vuong, Gupta, Gottardi, Norelli, Venkatesh, Gholamidavoodi, Tabassum, Menezes, Kirubarajan, Mullokandov, Sabharwal, Herrick, Efrat, Erdem, Karakacs, Roberts, Loe, Zoph, Bojanowski, Ozyurt, Hedayatnia, Neyshabur, Inden, Stein, Ekmekci, Lin, Howald, Orinion, Diao, Dour, Stinson, Argueta, Ram'irez, Singh, Rathkopf, Meng, Baral, Wu, Callison-Burch, Waites, Voigt, Manning, Potts, Ramirez, Rivera, Siro, Raffel, Ashcraft, Garbacea, Sileo, Garrette, Hendrycks, Kilman, Roth, Freeman, Khashabi, Levy, Gonz'alez, Perszyk, Hernandez, Chen, Ippolito, Gilboa, Dohan, Drakard, Jurgens, Datta, Ganguli, Emelin, Kleyko, Yuret, Chen, Tam, Hupkes, Misra, Buzan, Mollo, Yang, Lee, Schrader, Shutova,
  Cubuk, Segal, Hagerman, Barnes, Donoway, Pavlick, Rodol{\`a}, Lam, Chu, Tang, Erdem, Chang, Chi, Dyer, Jerzak, Kim, Manyasi, Zheltonozhskii, Xia, Siar, Mart'inez-Plumed, Happ'e, Chollet, Rong, Mishra, Winata, de~Melo, Kruszewski, Parascandolo, Mariani, Wang, Jaimovitch-L'opez, Betz, Gur-Ari, Galijasevic, Kim, Rashkin, Hajishirzi, Mehta, Bogar, Shevlin, Schutze, Yakura, Zhang, Wong, Ng, Noble, Jumelet, Geissinger, Kernion, Hilton, Lee, Fisac, Simon, Koppel, Zheng, Zou, Koco'n, Thompson, Wingfield, Kaplan, Radom, Sohl-Dickstein, Phang, Wei, Yosinski, Novikova, Bosscher, Marsh, Kim, Taal, Engel, Alabi, Xu, Song, Tang, Waweru, Burden, Miller, Balis, Batchelder, Berant, Frohberg, Rozen, Hern{\'a}ndez-Orallo, Boudeman, Guerr, Jones, Tenenbaum, Rule, Chua, Kanclerz, Livescu, Krauth, Gopalakrishnan, Ignatyeva, Markert, Dhole, Gimpel, Omondi, Mathewson, Chiafullo, Shkaruta, Shridhar, McDonell, Richardson, Reynolds, Gao, Zhang, Dugan, Qin, Contreras-Ochando, Morency, Moschella, Lam, Noble, Schmidt, He, Col'on, Metz,
  cSenel, Bosma, Sap, ter Hoeve, Farooqi, Faruqui, Mazeika, Baturan, Marelli, Maru, Quintana, Tolkiehn, Giulianelli, Lewis, Potthast, Leavitt, Hagen, Schubert, Baitemirova, Arnaud, McElrath, Yee, Cohen, Gu, Ivanitskiy, Starritt, Strube, Swkedrowski, Bevilacqua, Yasunaga, Kale, Cain, Xu, Suzgun, Walker, Tiwari, Bansal, Aminnaseri, Geva, Gheini, MukundVarma, Peng, Chi, Lee, Krakover, Cameron, Roberts, Doiron, Martinez, Nangia, Deckers, Muennighoff, Keskar, Iyer, Constant, Fiedel, Wen, Zhang, Agha, Elbaghdadi, Levy, Evans, Casares, Doshi, Fung, Liang, Vicol, Alipoormolabashi, Liao, Liang, Chang, Eckersley, Htut, Hwang, Milkowski, Patil, Pezeshkpour, Oli, Mei, Lyu, Chen, Banjade, Rudolph, Gabriel, Habacker, Risco, Milliere, Garg, Barnes, Saurous, Arakawa, Raymaekers, Frank, Sikand, Novak, Sitelew, Lebras, Liu, Jacobs, Zhang, Salakhutdinov, Chi, Lee, Stovall, Teehan, Yang, Singh, Mohammad, Anand, Dillavou, Shleifer, Wiseman, Gruetter, Bowman, Schoenholz, Han, Kwatra, Rous, Ghazarian, Ghosh, Casey, Bischoff,
  Gehrmann, Schuster, Sadeghi, Hamdan, Zhou, Srivastava, Shi, Singh, Asaadi, Gu, Pachchigar, Toshniwal, Upadhyay, Debnath, Shakeri, Thormeyer, Melzi, Reddy, Makini, Lee, Torene, Hatwar, Dehaene, Divic, Ermon, Biderman, Lin, Prasad, Piantadosi, Shieber, Misherghi, Kiritchenko, Mishra, Linzen, Schuster, Li, Yu, Ali, Hashimoto, Wu, Desbordes, Rothschild, Phan, Wang, Nkinyili, Schick, Kornev, Tunduny, Gerstenberg, Chang, Neeraj, Khot, Shultz, Shaham, Misra, Demberg, Nyamai, Raunak, Ramasesh, Prabhu, Padmakumar, Srikumar, Fedus, Saunders, Zhang, Vossen, Ren, Tong, Zhao, Wu, Shen, Yaghoobzadeh, Lakretz, Song, Bahri, Choi, Yang, Hao, Chen, Belinkov, Hou, Hou, Bai, Seid, Zhao, Wang, Wang, Wang, and Wu]{Srivastava2022BeyondTI}
Aarohi Srivastava, Abhinav Rastogi, Abhishek Rao, Abu Awal~Md Shoeb, Abubakar Abid, Adam Fisch, Adam~R. Brown, Adam Santoro, Aditya Gupta, Adri{\`a} Garriga-Alonso, Agnieszka Kluska, Aitor Lewkowycz, Akshat Agarwal, Alethea Power, Alex Ray, Alex Warstadt, Alexander~W. Kocurek, Ali Safaya, Ali Tazarv, Alice Xiang, Alicia Parrish, Allen Nie, Aman Hussain, Amanda Askell, Amanda Dsouza, Ambrose Slone, Ameet~Annasaheb Rahane, Anantharaman~S. Iyer, Anders Andreassen, Andrea Madotto, Andrea Santilli, Andreas Stuhlmuller, Andrew~M. Dai, Andrew La, Andrew~Kyle Lampinen, Andy Zou, Angela Jiang, Angelica Chen, Anh Vuong, Animesh Gupta, Anna Gottardi, Antonio Norelli, Anu Venkatesh, Arash Gholamidavoodi, Arfa Tabassum, Arul Menezes, Arun Kirubarajan, Asher Mullokandov, Ashish Sabharwal, Austin Herrick, Avia Efrat, Aykut Erdem, Ayla Karakacs, B.~Ryan Roberts, Bao~Sheng Loe, Barret Zoph, Bartlomiej Bojanowski, Batuhan Ozyurt, Behnam Hedayatnia, Behnam Neyshabur, Benjamin Inden, Benno Stein, Berk Ekmekci, Bill~Yuchen Lin,
  Blake~Stephen Howald, Bryan Orinion, Cameron Diao, Cameron Dour, Catherine Stinson, Cedrick Argueta, C'esar~Ferri Ram'irez, Chandan Singh, Charles Rathkopf, Chenlin Meng, Chitta Baral, Chiyu Wu, Chris Callison-Burch, Chris Waites, Christian Voigt, Christopher~D. Manning, Christopher Potts, Cindy Ramirez, Clara~E. Rivera, Clemencia Siro, Colin Raffel, Courtney Ashcraft, Cristina Garbacea, Damien Sileo, Daniel~H Garrette, Dan Hendrycks, Dan Kilman, Dan Roth, Daniel Freeman, Daniel Khashabi, Daniel Levy, Daniel~Mosegu'i Gonz'alez, Danielle~R. Perszyk, Danny Hernandez, Danqi Chen, Daphne Ippolito, Dar Gilboa, David Dohan, David Drakard, David Jurgens, Debajyoti Datta, Deep Ganguli, Denis Emelin, Denis Kleyko, Deniz Yuret, Derek Chen, Derek Tam, Dieuwke Hupkes, Diganta Misra, Dilyar Buzan, Dimitri~Coelho Mollo, Diyi Yang, Dong-Ho Lee, Dylan Schrader, Ekaterina Shutova, Ekin~Dogus Cubuk, Elad Segal, Eleanor Hagerman, Elizabeth Barnes, Elizabeth~P. Donoway, Ellie Pavlick, Emanuele Rodol{\`a}, Emma Lam, Eric Chu,
  Eric Tang, Erkut Erdem, Ernie Chang, Ethan~A. Chi, Ethan Dyer, Ethan~J. Jerzak, Ethan Kim, Eunice~Engefu Manyasi, Evgenii Zheltonozhskii, Fanyue Xia, Fatemeh Siar, Fernando Mart'inez-Plumed, Francesca Happ'e, François Chollet, Frieda Rong, Gaurav Mishra, Genta~Indra Winata, Gerard de~Melo, Germ{\'a}n Kruszewski, Giambattista Parascandolo, Giorgio Mariani, Gloria Wang, Gonzalo Jaimovitch-L'opez, Gregor Betz, Guy Gur-Ari, Hana Galijasevic, Hannah Kim, Hannah Rashkin, Hannaneh Hajishirzi, Harsh Mehta, Hayden Bogar, Henry Shevlin, Hinrich Schutze, Hiromu Yakura, Hongming Zhang, Hugh~Mee Wong, Ian Ng, Isaac Noble, Jaap Jumelet, Jack Geissinger, John Kernion, Jacob Hilton, Jaehoon Lee, Jaime~Fern{\'a}ndez Fisac, James~B. Simon, James Koppel, James Zheng, James Zou, Jan Koco'n, Jana Thompson, Janelle Wingfield, Jared Kaplan, Jarema Radom, Jascha~Narain Sohl-Dickstein, Jason Phang, Jason Wei, Jason Yosinski, Jekaterina Novikova, Jelle Bosscher, Jennifer Marsh, Jeremy Kim, Jeroen Taal, Jesse Engel,
  Jesujoba~Oluwadara Alabi, Jiacheng Xu, Jiaming Song, Jillian Tang, Jane~W Waweru, John Burden, John Miller, John~U. Balis, Jonathan Batchelder, Jonathan Berant, Jorg Frohberg, Jos Rozen, Jos{\'e} Hern{\'a}ndez-Orallo, Joseph Boudeman, Joseph Guerr, Joseph Jones, Joshua Tenenbaum, Joshua~S. Rule, Joyce Chua, Kamil Kanclerz, Karen Livescu, Karl Krauth, Karthik Gopalakrishnan, Katerina Ignatyeva, Katja Markert, Kaustubh~D. Dhole, Kevin Gimpel, Kevin Omondi, Kory~Wallace Mathewson, Kristen Chiafullo, Ksenia Shkaruta, Kumar Shridhar, Kyle McDonell, Kyle Richardson, Laria Reynolds, Leo Gao, Li~Zhang, Liam Dugan, Lianhui Qin, Lidia Contreras-Ochando, Louis-Philippe Morency, Luca Moschella, Luca Lam, Lucy Noble, Ludwig Schmidt, Luheng He, Luis~Oliveros Col'on, Luke Metz, Lutfi~Kerem cSenel, Maarten Bosma, Maarten Sap, Maartje ter Hoeve, Maheen Farooqi, Manaal Faruqui, Mantas Mazeika, Marco Baturan, Marco Marelli, Marco Maru, Maria Jose~Ram'irez Quintana, Marie Tolkiehn, Mario Giulianelli, Martha Lewis, Martin
  Potthast, Matthew~L. Leavitt, Matthias Hagen, M'aty'as Schubert, Medina Baitemirova, Melody Arnaud, Melvin~Andrew McElrath, Michael~A. Yee, Michael Cohen, Michael Gu, Michael Ivanitskiy, Michael Starritt, Michael Strube, Michal Swkedrowski, Michele Bevilacqua, Michihiro Yasunaga, Mihir Kale, Mike Cain, Mimee Xu, Mirac Suzgun, Mitch Walker, Monica Tiwari, Mohit Bansal, Moin Aminnaseri, Mor Geva, Mozhdeh Gheini, T~MukundVarma, Nanyun Peng, Nathan~A. Chi, Nayeon Lee, Neta Gur-Ari Krakover, Nicholas Cameron, Nicholas Roberts, Nick Doiron, Nicole Martinez, Nikita Nangia, Niklas Deckers, Niklas Muennighoff, Nitish~Shirish Keskar, Niveditha Iyer, Noah Constant, Noah Fiedel, Nuan Wen, Oliver Zhang, Omar Agha, Omar Elbaghdadi, Omer Levy, Owain Evans, Pablo Antonio~Moreno Casares, Parth Doshi, Pascale Fung, Paul~Pu Liang, Paul Vicol, Pegah Alipoormolabashi, Peiyuan Liao, Percy Liang, Peter Chang, Peter Eckersley, Phu~Mon Htut, Pi-Bei Hwang, P.~Milkowski, Piyush~S. Patil, Pouya Pezeshkpour, Priti Oli, Qiaozhu Mei,
  Qing Lyu, Qinlang Chen, Rabin Banjade, Rachel~Etta Rudolph, Raefer Gabriel, Rahel Habacker, Ramon Risco, Raphael Milliere, Rhythm Garg, Richard Barnes, Rif~A. Saurous, Riku Arakawa, Robbe Raymaekers, Robert Frank, Rohan Sikand, Roman Novak, Roman Sitelew, Ronan Lebras, Rosanne Liu, Rowan Jacobs, Rui Zhang, Ruslan Salakhutdinov, Ryan Chi, Ryan Lee, Ryan Stovall, Ryan Teehan, Rylan Yang, Sahib Singh, Saif~M. Mohammad, Sajant Anand, Sam Dillavou, Sam Shleifer, Sam Wiseman, Samuel Gruetter, Samuel~R. Bowman, Samuel~S. Schoenholz, Sanghyun Han, Sanjeev Kwatra, Sarah~A. Rous, Sarik Ghazarian, Sayan Ghosh, Sean Casey, Sebastian Bischoff, Sebastian Gehrmann, Sebastian Schuster, Sepideh Sadeghi, Shadi~S. Hamdan, Sharon Zhou, Shashank Srivastava, Sherry Shi, Shikhar Singh, Shima Asaadi, Shixiang~Shane Gu, Shubh Pachchigar, Shubham Toshniwal, Shyam Upadhyay, Shyamolima Debnath, Siamak Shakeri, Simon Thormeyer, Simone Melzi, Siva Reddy, Sneha~Priscilla Makini, Soo-Hwan Lee, Spencer Torene, Sriharsha Hatwar, Stanislas
  Dehaene, Stefan Divic, Stefano Ermon, Stella Biderman, Stephanie Lin, Stephen Prasad, Steven~T Piantadosi, Stuart~M. Shieber, Summer Misherghi, Svetlana Kiritchenko, Swaroop Mishra, Tal Linzen, Tal Schuster, Tao Li, Tao Yu, Tariq Ali, Tatsunori Hashimoto, Te-Lin Wu, Theo Desbordes, Theodore Rothschild, Thomas Phan, Tianle Wang, Tiberius Nkinyili, Timo Schick, Timofei Kornev, Titus Tunduny, Tobias Gerstenberg, Trenton Chang, Trishala Neeraj, Tushar Khot, Tyler Shultz, Uri Shaham, Vedant Misra, Vera Demberg, Victoria Nyamai, Vikas Raunak, Vinay~Venkatesh Ramasesh, Vinay~Uday Prabhu, Vishakh Padmakumar, Vivek Srikumar, William Fedus, William Saunders, William Zhang, Wout Vossen, Xiang Ren, Xiaoyu Tong, Xinran Zhao, Xinyi Wu, Xudong Shen, Yadollah Yaghoobzadeh, Yair Lakretz, Yangqiu Song, Yasaman Bahri, Yejin Choi, Yichi Yang, Yiding Hao, Yifu Chen, Yonatan Belinkov, Yu~Hou, Yu~Hou, Yuntao Bai, Zachary Seid, Zhuoye Zhao, Zi~Fu Wang, Zijie~J. Wang, Zirui Wang, and Ziyi Wu.
\newblock Beyond the imitation game: Quantifying and extrapolating the capabilities of language models.
\newblock 2022.
\newblock URL \url{https://api.semanticscholar.org/CorpusID:263625818}.

\bibitem[Stiennon et~al.(2020)Stiennon, Ouyang, Wu, Ziegler, Lowe, Voss, Radford, Amodei, and Christiano]{Stiennon2020LearningTS}
Nisan Stiennon, Long Ouyang, Jeff Wu, Daniel~M. Ziegler, Ryan~J. Lowe, Chelsea Voss, Alec Radford, Dario Amodei, and Paul Christiano.
\newblock Learning to summarize from human feedback.
\newblock \emph{ArXiv}, abs/2009.01325, 2020.
\newblock URL \url{https://api.semanticscholar.org/CorpusID:221665105}.

\bibitem[Sun et~al.(2023)Sun, Gupta, and Iyyer]{Sun2023ExploringTI}
Simeng Sun, Dhawal Gupta, and Mohit Iyyer.
\newblock Exploring the impact of low-rank adaptation on the performance, efficiency, and regularization of rlhf.
\newblock \emph{ArXiv}, abs/2309.09055, 2023.
\newblock URL \url{https://api.semanticscholar.org/CorpusID:261884455}.

\bibitem[Touvron et~al.(2023)Touvron, Martin, Stone, Albert, Almahairi, Babaei, Bashlykov, Batra, Bhargava, Bhosale, Bikel, Blecher, Ferrer, Chen, Cucurull, Esiobu, Fernandes, Fu, Fu, Fuller, Gao, Goswami, Goyal, Hartshorn, Hosseini, Hou, Inan, Kardas, Kerkez, Khabsa, Kloumann, Korenev, Koura, Lachaux, Lavril, Lee, Liskovich, Lu, Mao, Martinet, Mihaylov, Mishra, Molybog, Nie, Poulton, Reizenstein, Rungta, Saladi, Schelten, Silva, Smith, Subramanian, Tan, Tang, Taylor, Williams, Kuan, Xu, Yan, Zarov, Zhang, Fan, Kambadur, Narang, Rodriguez, Stojnic, Edunov, and Scialom]{touvron2023llama}
Hugo Touvron, Louis Martin, Kevin Stone, Peter Albert, Amjad Almahairi, Yasmine Babaei, Nikolay Bashlykov, Soumya Batra, Prajjwal Bhargava, Shruti Bhosale, Dan Bikel, Lukas Blecher, Cristian~Canton Ferrer, Moya Chen, Guillem Cucurull, David Esiobu, Jude Fernandes, Jeremy Fu, Wenyin Fu, Brian Fuller, Cynthia Gao, Vedanuj Goswami, Naman Goyal, Anthony Hartshorn, Saghar Hosseini, Rui Hou, Hakan Inan, Marcin Kardas, Viktor Kerkez, Madian Khabsa, Isabel Kloumann, Artem Korenev, Punit~Singh Koura, Marie-Anne Lachaux, Thibaut Lavril, Jenya Lee, Diana Liskovich, Yinghai Lu, Yuning Mao, Xavier Martinet, Todor Mihaylov, Pushkar Mishra, Igor Molybog, Yixin Nie, Andrew Poulton, Jeremy Reizenstein, Rashi Rungta, Kalyan Saladi, Alan Schelten, Ruan Silva, Eric~Michael Smith, Ranjan Subramanian, Xiaoqing~Ellen Tan, Binh Tang, Ross Taylor, Adina Williams, Jian~Xiang Kuan, Puxin Xu, Zheng Yan, Iliyan Zarov, Yuchen Zhang, Angela Fan, Melanie Kambadur, Sharan Narang, Aurelien Rodriguez, Robert Stojnic, Sergey Edunov, and Thomas
  Scialom.
\newblock Llama 2: Open foundation and fine-tuned chat models, 2023.

\bibitem[Uesato et~al.(2022)Uesato, Kushman, Kumar, Song, Siegel, Wang, Creswell, Irving, and Higgins]{Uesato2022SolvingMW}
Jonathan Uesato, Nate Kushman, Ramana Kumar, Francis Song, Noah Siegel, L.~Wang, Antonia Creswell, Geoffrey Irving, and Irina Higgins.
\newblock Solving math word problems with process- and outcome-based feedback.
\newblock \emph{ArXiv}, abs/2211.14275, 2022.
\newblock URL \url{https://api.semanticscholar.org/CorpusID:254017497}.

\bibitem[Vinyals et~al.(2019)Vinyals, Babuschkin, Czarnecki, Mathieu, Dudzik, Chung, Choi, Powell, Ewalds, Georgiev, Oh, Horgan, Kroiss, Danihelka, Huang, Sifre, Cai, Agapiou, Jaderberg, Vezhnevets, Leblond, Pohlen, Dalibard, Budden, Sulsky, Molloy, Paine, Gulcehre, Wang, Pfaff, Wu, Ring, Yogatama, W{\"u}nsch, McKinney, Smith, Schaul, Lillicrap, Kavukcuoglu, Hassabis, Apps, and Silver]{Vinyals2019GrandmasterLI}
Oriol Vinyals, Igor Babuschkin, Wojciech~M. Czarnecki, Micha{\"e}l Mathieu, Andrew Dudzik, Junyoung Chung, David~H. Choi, Richard Powell, Timo Ewalds, Petko Georgiev, Junhyuk Oh, Dan Horgan, Manuel Kroiss, Ivo Danihelka, Aja Huang, L.~Sifre, Trevor Cai, John~P. Agapiou, Max Jaderberg, Alexander~Sasha Vezhnevets, R{\'e}mi Leblond, Tobias Pohlen, Valentin Dalibard, David Budden, Yury Sulsky, James Molloy, Tom~Le Paine, Caglar Gulcehre, Ziyun Wang, Tobias Pfaff, Yuhuai Wu, Roman Ring, Dani Yogatama, Dario W{\"u}nsch, Katrina McKinney, Oliver Smith, Tom Schaul, Timothy~P. Lillicrap, Koray Kavukcuoglu, Demis Hassabis, Chris Apps, and David Silver.
\newblock Grandmaster level in starcraft ii using multi-agent reinforcement learning.
\newblock \emph{Nature}, 575:\penalty0 350 -- 354, 2019.
\newblock URL \url{https://api.semanticscholar.org/CorpusID:204972004}.

\bibitem[Wei et~al.(2021)Wei, Bosma, Zhao, Guu, Yu, Lester, Du, Dai, and Le]{Wei2021FinetunedLM}
Jason Wei, Maarten Bosma, Vincent Zhao, Kelvin Guu, Adams~Wei Yu, Brian Lester, Nan Du, Andrew~M. Dai, and Quoc~V. Le.
\newblock Finetuned language models are zero-shot learners.
\newblock \emph{ArXiv}, abs/2109.01652, 2021.
\newblock URL \url{https://api.semanticscholar.org/CorpusID:237416585}.

\bibitem[Wei et~al.(2022)Wei, Wang, Schuurmans, Bosma, hsin Chi, Xia, Le, and Zhou]{Wei2022ChainOT}
Jason Wei, Xuezhi Wang, Dale Schuurmans, Maarten Bosma, Ed~Huai hsin Chi, F.~Xia, Quoc Le, and Denny Zhou.
\newblock Chain of thought prompting elicits reasoning in large language models.
\newblock \emph{ArXiv}, abs/2201.11903, 2022.

\bibitem[Yao et~al.(2020)Yao, Rao, Hausknecht, and Narasimhan]{Yao2020KeepCA}
Shunyu Yao, Rohan Rao, Matthew~J. Hausknecht, and Karthik Narasimhan.
\newblock Keep calm and explore: Language models for action generation in text-based games.
\newblock \emph{ArXiv}, abs/2010.02903, 2020.
\newblock URL \url{https://api.semanticscholar.org/CorpusID:222142129}.

\bibitem[Yao et~al.(2022)Yao, Zhao, Yu, Du, Shafran, Narasimhan, and Cao]{Yao2022ReActSR}
Shunyu Yao, Jeffrey Zhao, Dian Yu, Nan Du, Izhak Shafran, Karthik Narasimhan, and Yuan Cao.
\newblock React: Synergizing reasoning and acting in language models.
\newblock \emph{ArXiv}, abs/2210.03629, 2022.
\newblock URL \url{https://api.semanticscholar.org/CorpusID:252762395}.

\bibitem[Yao et~al.(2023)Yao, Yu, Zhao, Shafran, Griffiths, Cao, and Narasimhan]{Yao2023TreeOT}
Shunyu Yao, Dian Yu, Jeffrey Zhao, Izhak Shafran, Thomas~L. Griffiths, Yuan Cao, and Karthik Narasimhan.
\newblock Tree of thoughts: Deliberate problem solving with large language models.
\newblock \emph{ArXiv}, abs/2305.10601, 2023.
\newblock URL \url{https://api.semanticscholar.org/CorpusID:258762525}.

\bibitem[Ye et~al.(2022)Ye, Cui, Shi, and Riedl]{Ye2022NeuralSP}
Anbang Ye, Christopher Cui, Taiwei Shi, and Mark~O. Riedl.
\newblock Neural story planning.
\newblock \emph{ArXiv}, abs/2212.08718, 2022.
\newblock URL \url{https://api.semanticscholar.org/CorpusID:254854533}.

\bibitem[Yuan et~al.(2023)Yuan, Yuan, Li, Dong, Tan, and Zhou]{Yuan2023ScalingRO}
Zheng Yuan, Hongyi Yuan, Cheng Li, Guanting Dong, Chuanqi Tan, and Chang Zhou.
\newblock Scaling relationship on learning mathematical reasoning with large language models.
\newblock \emph{ArXiv}, abs/2308.01825, 2023.
\newblock URL \url{https://api.semanticscholar.org/CorpusID:260438790}.

\bibitem[Zelikman et~al.(2022)Zelikman, Wu, Mu, and Goodman]{zelikman2022star}
Eric Zelikman, Yuhuai Wu, Jesse Mu, and Noah~D. Goodman.
\newblock Star: Bootstrapping reasoning with reasoning, 2022.

\bibitem[Zhou et~al.(2023{\natexlab{a}})Zhou, Wang, Lu, Shi, Luo, Qin, Lu, Jia, Song, Zhan, and Li]{Zhou2023SolvingCM}
Aojun Zhou, Ke~Wang, Zimu Lu, Weikang Shi, Sichun Luo, Zipeng Qin, Shaoqing Lu, Anya Jia, Linqi Song, Mingjie Zhan, and Hongsheng Li.
\newblock Solving challenging math word problems using gpt-4 code interpreter with code-based self-verification.
\newblock \emph{ArXiv}, abs/2308.07921, 2023{\natexlab{a}}.
\newblock URL \url{https://api.semanticscholar.org/CorpusID:260900008}.

\bibitem[Zhou et~al.(2023{\natexlab{b}})Zhou, Peng, and Riedl]{Zhou2023DialogueSE}
Wei Zhou, Xiangyu Peng, and Mark~O. Riedl.
\newblock Dialogue shaping: Empowering agents through npc interaction.
\newblock \emph{ArXiv}, abs/2307.15833, 2023{\natexlab{b}}.
\newblock URL \url{https://api.semanticscholar.org/CorpusID:260333931}.

\bibitem[Zhu et~al.(2023)Zhu, Wang, Zhang, Zhang, Huang, Gan, Zhang, and Yang]{Zhu_2023}
Xinyu Zhu, Junjie Wang, Lin Zhang, Yuxiang Zhang, Yongfeng Huang, Ruyi Gan, Jiaxing Zhang, and Yujiu Yang.
\newblock Solving math word problems via cooperative reasoning induced language models.
\newblock Association for Computational Linguistics, 2023.
\newblock \doi{10.18653/v1/2023.acl-long.245}.
\newblock URL \url{https://doi.org/10.18653%2Fv1%2F2023.acl-long.245}.

\bibitem[Ziegler et~al.(2019)Ziegler, Stiennon, Wu, Brown, Radford, Amodei, Christiano, and Irving]{ziegler2019fine}
Daniel~M Ziegler, Nisan Stiennon, Jeffrey Wu, Tom~B Brown, Alec Radford, Dario Amodei, Paul Christiano, and Geoffrey Irving.
\newblock Fine-tuning language models from human preferences.
\newblock \emph{arXiv preprint arXiv:1909.08593}, 2019.

\end{thebibliography}

\newpage

\appendix

\section{RCRL Label Balance}
\label{sec:rcrl_label_balance}

We also experiment with different proportions of `[GOOD]' and `[BAD]' labels in RCRL training data. This is motivated by a desire to make better use of abundant negative data, which is much easier to generate than its positive counterpart. Better teaching the student what \textbf{not} to do with this data would ideally increase the number of valid solutions. Recall by default we balance the number of positive and negative samples.

We conduct experiments on LLama-2 7 GSM8K without any SFT data. We apply only one round of Expert Iteration ($K=1$ per question), producing a student model we refer to as \textbf{EI-minimal}. Note, in this setting we only provide `[GOOD]' and `[BAD]' labels for entire solutions, rather than providing labels at the step level. Results are reported in \ref{tab:csft_proportion}.

\begin{table}[h]
    \centering
    \begin{tabular}{c|c|cc|cc} \toprule
        & positive:negative ratio &   GSM8K (maj@1)\\ \hline 
        \textbf{EI-minimal} & - & 0.17 \\ \hline 
        & 100:1 & \textbf{0.18}\\  
       \textbf{RCRL} & 10:1 & \textbf{0.18}\\ 
        & 1:1 &  0.15\\ \bottomrule
    \end{tabular}
    \caption{\textbf{RCRL} without SFT, using different proportions of positive and negative samples. As we increase the proportion of negative samples, performance generally decreases. At best, we only see very marginal gains using \textbf{RCRL}. Note: \textbf{EI-minimal} refers to running EI for one iteration, with $K=1$ per question.}
    \label{tab:csft_proportion}
\end{table}

We find we achieve best performance when the amount of positive training data greatly outweighs the amount of negative data. In these cases, our RCRL models' maj@1 score slightly exceeds the maj@1 score of the data generating EI-minimal model. Yet, when we balance the amount of positive and negative training data, we find performance is degraded. This suggests our 7B student doesn't effectively learn from the provided negative demonstrations. We suspect either a larger model or an easier task would give better results.

\section{EI Improvement across Iterations}
\label{sec:diversity}

\begin{figure}
    \begin{minipage}[ht]{0.48\textwidth}
        \centering
        \includegraphics[scale=0.5]{figs/gsm8k_ei_round_test_acc.png}
        \caption{Accuracy of EI models on GSM8K test vs. number of iterations. \textit{EI scratch} models have use no SFT initialization.}
        \label{fig:gsm8k_ei_round_acc_appendix}
    \end{minipage}
    \begin{minipage}[ht]{0.48\textwidth}
        \centering
        \includegraphics[scale=0.5]{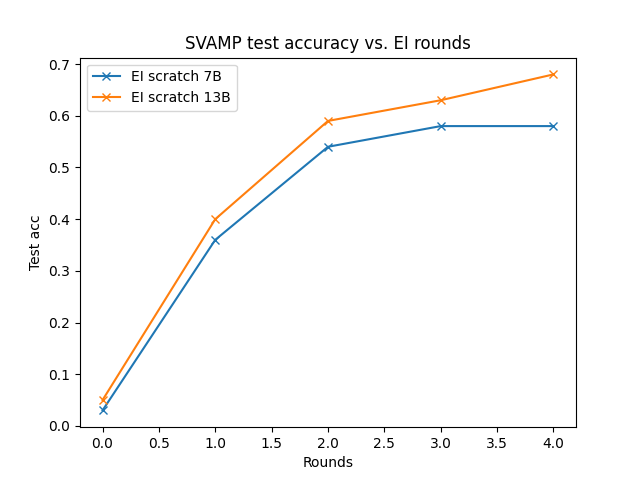}
        \caption{Accuracy of EI models on GSM8K test vs. number of iterations. $K = 4$ samples per prompt are used to construct a fine-tuning dataset for the next round.}
        \label{fig:svamp_ei_round_acc}
    \end{minipage}
\end{figure}

Figures \ref{fig:gsm8k_ei_round_acc_appendix} and \ref{fig:svamp_ei_round_acc} plot the maj@1 score of models on versus rounds of expert iteration. On both datasets the score is monotonically increasing until convergence after at most four rounds. Models initialized from an SFT checkpoint converge faster than their pretrained counterparts. Each round of expert iteration samples $K * \texttt{num\_train}$ rollouts, with the longest running training loop generate at most $5 * 4 * 7000 \approx 10^6$ samples.

\begin{figure}
    \centering
    \includegraphics[scale=0.5]{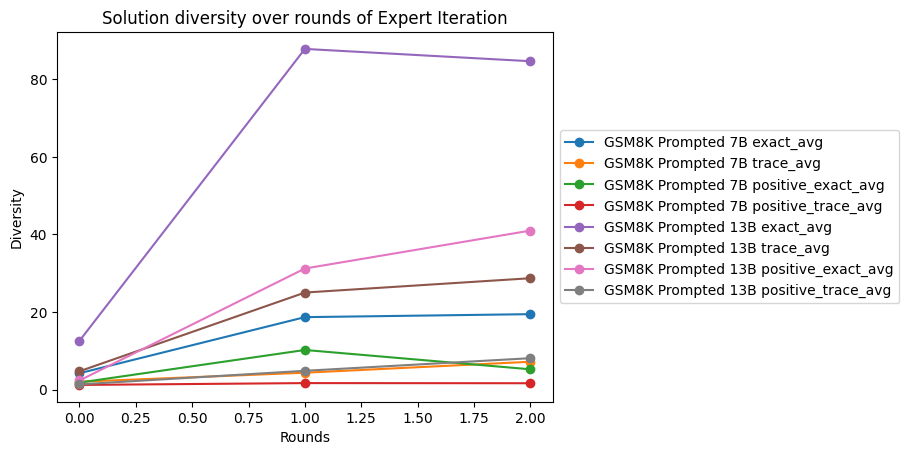}
    \caption{Diversity of GSM8K model output over rounds of EI. (No SFT)}
    \label{fig:gsm8k_ei_prompted_diversity}
\end{figure}

\begin{figure}
        \centering
        \includegraphics[scale=0.5]{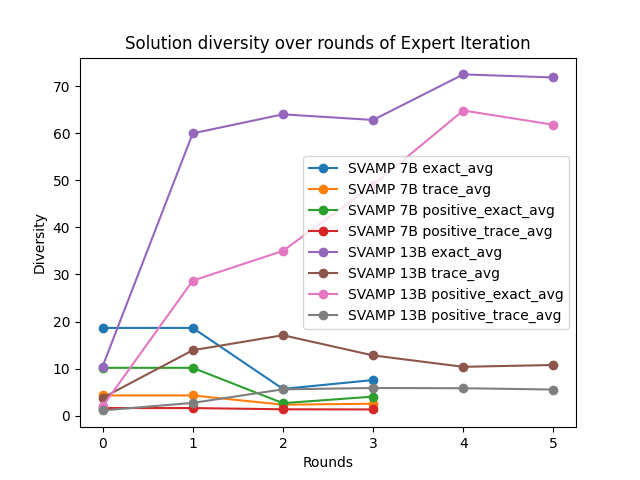}
        \caption{Diversity of SVAMP model output over rounds of EI. $K = 96$ samples are used per prompt. \textit{positive} diversity measures diversity in the subset of solutions with a correct final answer.}
    \label{fig:svamp_ei_diversity}
\end{figure}

Figures \ref{fig:gsm8k_ei_prompted_diversity} and \ref{fig:svamp_ei_diversity} report the diversity of solutions across rounds of expert iteration as measured by two separate metrics for solution uniqueness. \textit{exact diversity} checks for equality between two solutions using exact string match. \textit{trace diversity} checks for equality between two solutions by first extracting the \textit{trace} of a solution as the sequence of intermediate calculations used to get to the final answer. An exact match is then performed on this trace representation. 

\textbf{Solution diversity increases then decreases over training} Overall both measures of solution diversity increase for both model sizes over the first two rounds of expert iteration. After the first two rounds both trace diversity appears to plateau and in some cases slightly decrease. Exact diversity continues to increase for 13B, but not at the same rate as during the first two rounds. The largest increases in solution diversity over the first two rounds also match when the largest gains in maj@1 performance occur. This lends evidence to the intuition that a high-performing student will be able to generate many correct but unique solutionst to the same problem. Further, we see during later rounds of expert iteration that while maj@1 score improves slightly, diversity suffers. This provides further evidence that training is begining to overfit to maj@1 score, in the process reducing both pass@n and solution diversity. We see the same behavior 

\begin{figure}
    \centering
    \includegraphics[scale=0.5]{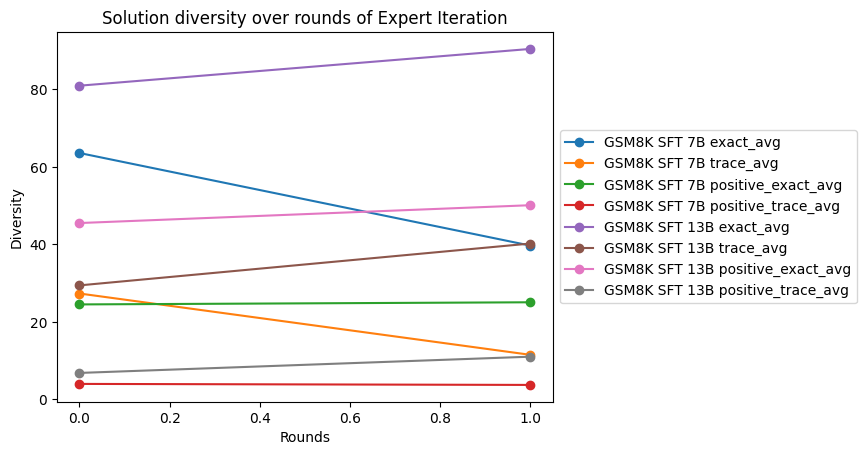}
    \caption{gsm8k sft diversity}
    \label{fig:gsm8k_ei_sft_diversity}
\end{figure}

\textbf{Larger models generate more diverse solutions} The above figures also demonstrate the 13B model produces signifcantly more diverse outputs than the 7B model. This is true during every round of fine-tuning, with the gap getting larger as more training is done. Interestingly, the 13B model appears to produce an \textit{exactly} unique solution with every sampling after 4 rounds of expert iteration. However, its trace diversity peaks after two rounds, indicating 13B tends to introduce semantic diversity without changing the underlying computational structure of a solution.

\section{Sample Complexities}

\begin{figure}[ht]
  \begin{minipage}[t]{0.48\textwidth} 
    \centering
    \includegraphics[scale=0.5]{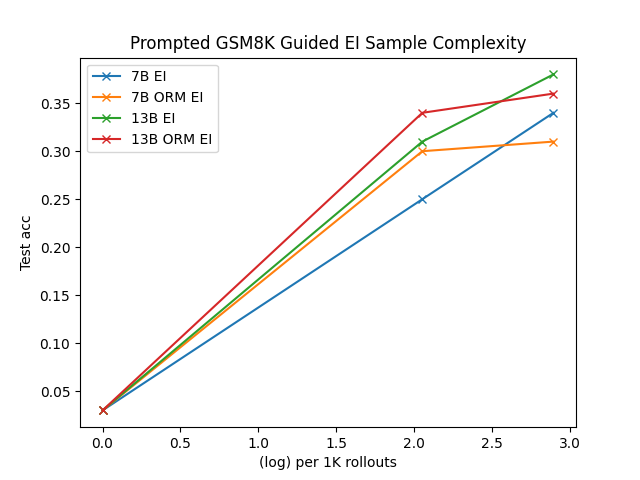}
    \caption{Sample complexity of default versus ORM guided EI students on GSM8K (no SFT). The ORM improves sample complexity initially but ultimately underperforms using only the ground truth.}
    \label{fig:prompted-gsm8k-guided-ei-sample-complexity}
  \end{minipage}
  \hfill 
  \begin{minipage}[t]{0.48\textwidth} 
    \centering
    \includegraphics[scale=0.5]{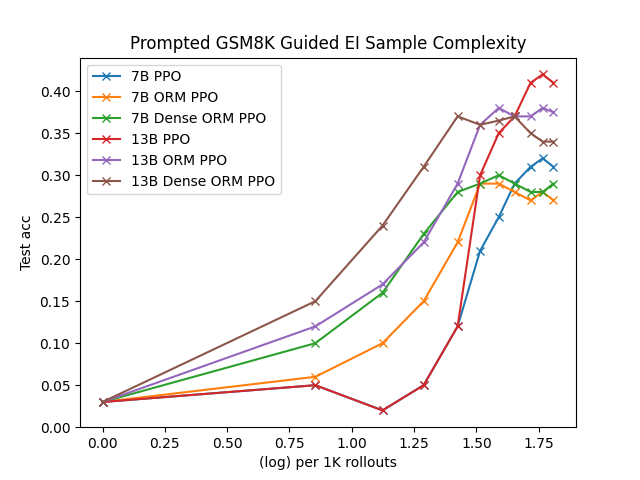}
    \caption{Sample complexity of default versus ORM guided PPO students on GSM8K (no SFT). Similarly to as in EI, the ORM improves maj@1 score over using only ground truth rewards but eventually underperforms.}
    \label{fig:prompted-gsm8k-guided-ppo-sample-complexity}
  \end{minipage}
\end{figure}

\begin{figure}[ht]
  \begin{minipage}[t]{0.48\textwidth} 
     \centering
    \includegraphics[scale=0.5]{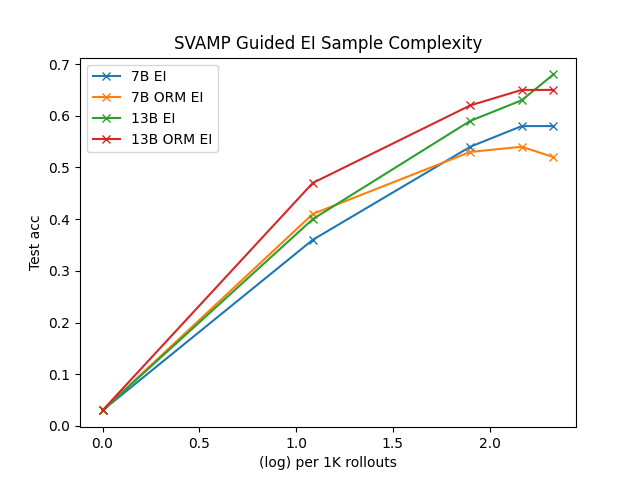}
    \caption{Sample complexity of default versus ORM guided EI students on SVAMP.}
    \label{fig:svamp-guided-ei-sample-complexity}
  \end{minipage}
  \hfill 
  \begin{minipage}[t]{0.48\textwidth} 
    \centering
    \includegraphics[scale=0.5]{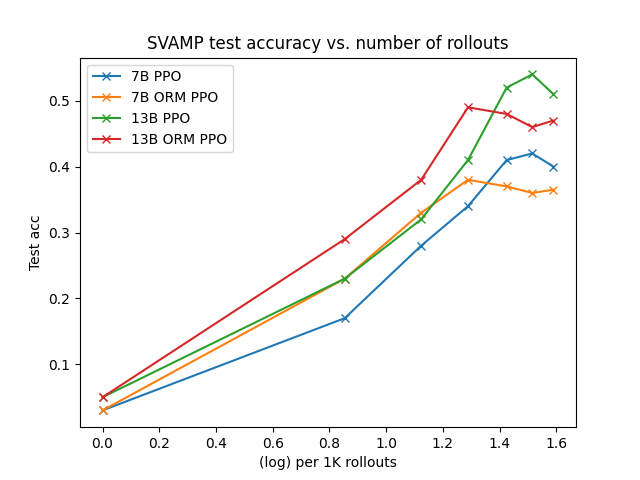}
    \caption{Sample complexity of default versus ORM guided PPO students on SVAMP.}
    \label{fig:svamp-guided-ppo-sample-complexity}
  \end{minipage}
\end{figure}

In this section we plot all sample complexities on benchmarks accompanying the results in Section \ref{sec:experiments}. Figures \ref{fig:prompted-gsm8k-guided-ei-sample-complexity} and \ref{fig:prompted-gsm8k-guided-ppo-sample-complexity} report results on GSM8K without supervised fine-tuning. Figures \ref{fig:svamp-guided-ei-sample-complexity} and \ref{fig:svamp-guided-ppo-sample-complexity} report results on SVAMP.

As in the SFT case, using an ORM to guide EI and PPO on prompted GSM8K models does somewhat reduce sample complexity but does not improve best performance (if anything the ORM reward slightly hurts converged maj@1 score). We see the same story when providing a dense ORM reward, further decreasing sample comlexity but at the cost of final converged performance. Our best results still come from using only the ground truth score. We suspect the performance degredation introduced by the ORM reward could be alleviated with a larger reward model. However, we do not believe using a larger model would improve over just the ground truth reward. Similar results are seen for SVAMP.

\section{Curriculum Learning for RL}

\begin{figure}
    \centering
    \includegraphics[scale=0.5]{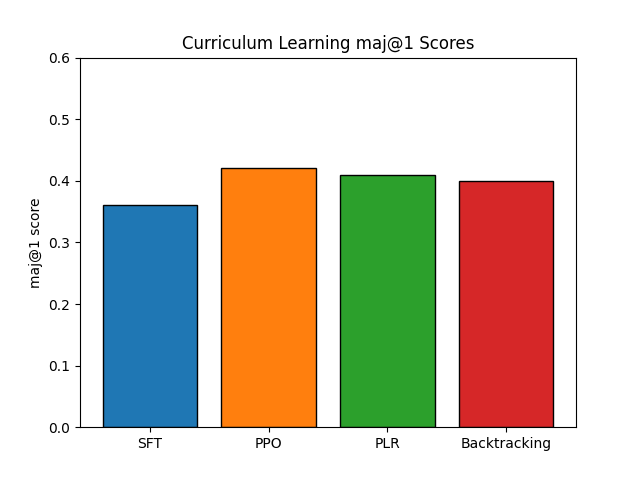}
    \caption{maj@1 scores for Prioritized Level Replay (PLR) and Backtracking techniques compared to default PPO and SFT.}
    \label{fig:curriculum_results}
\end{figure}

In addition to vanilla PPO we experiment with backtracking \citep{Salimans2018LearningMR} and Prioritized Level Replay (\textbf{PLR}) \citep{Jiang2020PrioritizedLR} as algorithms from the curriculum learning literature. Such algorithms aim to construct a ``curriculum'' of subproblems, with the model ideally learning to generalize from easier subproblems to harder subproblems. 

Backtracking in particular is a natural choice as it relies on using high-quality supervised trajectories to improve exploration of the solution space. This is done by sampling the student policy $\pi$ on the partially complete solution $(Q, P_i)$ where $P_i$ is a sequence of intermediate ground truth steps $(S_1,...,S_i)$. The algorithm proceeds by setting an initial threshold $\tau_0 \in (0, 1)$ which represents how far back from the final answer to initialize partial solutions. By default we use $\tau_0 = 0.9$. Then, for each problem $Q$ which can be solved from $P_i$, we remove the last step $S_i$ and condition on $P_{i-1}$ the next time $Q$ is sampled. 

PLR does not rely on access to SFT data, instead heuristically prioritizing problems with high ``learning potential'' estimated by the average absolute advantage. Prioritizing problems with this potential allows the model to focus on problems that are neither too easy nor too hard, making efficient use of its exploration budget. We initialize the student using a supervised fine-tuned LLama-2 7B on GSM8K. Results are reported in Figure \ref{fig:curriculum_results}.

\begin{figure}
    \centering
    \includegraphics[scale=0.5]{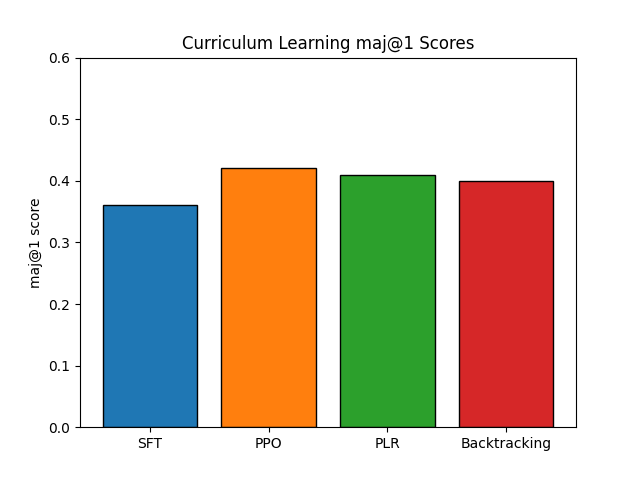}
    \caption{maj@1 scores on GSM8K for Prioritized Level Replay (PLR) and Backtracking techniques compared to default PPO and SFT.}
    \label{fig:curriculum_results}
\end{figure}

Overall we find neither method exceeds the performance of default PPO. We hypothesize this is due to the limited exploration the model engages in from the start, due to both pretraining and supervised fine-tuning. We speculate better results might be achieved on a harder dataset with more intermediate steps, particularly when using backtracking.

\section{Data augmentation}

We additionally experimented with generating synthetic $(Q, A)$ training pairs via an approach inspired by backtranslation \citep{Sennrich2015ImprovingNM}. We assume access to a supervised fine-tuning dataset $\mathcal{D}$ of $(Q, A)$ pairs and train a $Q\to A$ model $M_{Q\to A}$ as our usual student model. We call this model the verifier. We can also utilize $\mathcal{D}$ to train models of the form $M_{A \to Q}$ and $M_{A \to A}$ which map answers to questions and answers to answers respectively. We train $M_{A \to Q}$ simply by fine-tuning the pretrained model $M$ to predict $p(A|Q)$ where $(Q, A) \sim \mathcal{D}$. We call the combination of $M_{A \to A}$ and $M_{A \to Q}$ the generator. We construct a train set for $M_{A \to A}$ as follows: For each $A$ in $(Q, A) \in \mathcal{D}$ we randomly sample three other answers $A_1, A_2, A_3$ from $\mathcal{D}$ which act as a conditional prompt. We then train $M_{A \to A}$ by minimizing $p(A | A_1, A_2, A_3)$. 

\begin{figure}
    \centering
    \includegraphics[scale=0.75]{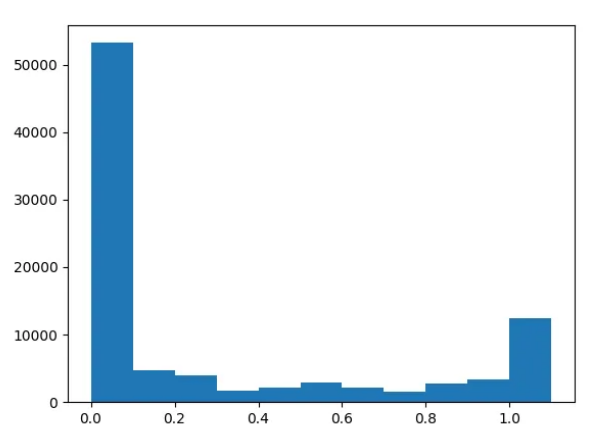}
    \caption{Scores of synthetically backwards generated $(Q, A)$ pairs. Note: the score refers to the percentage of times the forward student model $M_{Q \to A}$ recovers the intended final answer.}
    \label{fig:augmentation_results}
\end{figure}

We sample $M_{A \to A}$ on each ground truth answer $A \in \mathcal{D}$ $K = 8$ times, constructing a synthetic dataset of answers $\mathcal{A}$. We then use our backwards model $M_{A \to Q}$ to produce questions for each of the synthetic answers $A \in \mathcal{A}$. This forms a synthetic $(Q, A)$ dataset $\mathcal{D}_{\textup{synth}}$. Finally, for each synthetic $(Q, A)$ pair, we sample our student model $M_{Q \to A}$ $K = 20$ times for each question and check whether the student model's final answer agrees with the ``intended'' final answer. We refer to the percentage of student generated solutions recovering the intended final answer as the \textit{score} for a synthetic $(Q, A)$ pair. We plot the distribution of scores in Figure \ref{fig:augmentation_results}.

We see that the majority of synthetic pairs, over 50,000, never have their solutions recovered by the student $M_{Q\to A}$. This is either because a) the student is too weak to solve the question or b) the question is impossible to solve. Either way, we likely do not want to include these new training data for the student. Similarly, we likely do not want to include questions which are always solved by the student, i.e. those with score = 1, as they are too easy. Additionally, we should be wary of questions which have a small score in the range $(0, \epsilon)$. We expect many questions will have been solved incorrectly but still arriving at the correct final answer. We should exclude such problems from our training dataset.

We expect the highest quality data $(Q,A)$ to have a score in the neighborhood $(\frac{1}{2} - \tau, \frac{1}{2}+ \tau)$. These questions should be not too hard but not too easy for our student. Figure \ref{tab:model_perf_vs_tau} shows the performance of student models fine-tuned on a combination of ground truth data and synthetically generated data with scores in the range $(\frac{1}{2} - \tau, \frac{1}{2} + \tau)$. All models are trained for five epochs with an initial lr = 2e-5 cosine decayed to 2e-7. Llama-2 7B is used as the pretrained base model.

\begin{table}[t]
    \centering
    \begin{tabular}{@{}lr@{}}
          & maj@1 \\
        \toprule
         $\tau = 0.1$ & 0.38\\
         $\tau = 0.2$ & 0.36\\
         $\tau = 0.3$ & 0.34\\
         \midrule
         SFT & 0.41 \\
         \bottomrule
    \end{tabular}
    \caption{Performance of models training with various amounts of synthetic data vs. the SFT baseline. Note: $\tau$ represents the size of the neighborhood of scores around $\frac{1}{2}$ that are not filtered out.}
    \label{tab:model_perf_vs_tau}
\end{table}

Unfortunately, it seems introducing any amount of synthetically generated data degrades performance. When manually inspecting the synthetically generated $(Q, A)$ pairs it becomes clear why. There is an extremely high number of false positives.  Consider the following example of a synthetic pair shown in Table \ref{tab:synthetic_example}:

\begin{table}[t]
    \centering
    \begin{tabular}{@{}l|p{8cm}@{}}
        \toprule
        Question & "A school of 100 musicians goes on a skiing trip. 40\% are beginners, 30\% are intermediate, and 50\% are advanced. How many people went on the skiing trip?" \\
        \midrule
        Answer & "There are 100 * 0.4 = 40 beginner skiiers. There are 100 * 0.3 = 30 intermediate skiiers. There are 100 * 0.5 = 50 advanced skiiers. Therefore there are 40 + 30 + 50 = 120 skiiers total." \\
         \bottomrule
    \end{tabular}
    \label{tab:synthetic_example}
\end{table}

This is an example of a low-quality sample we do not want in our training data. Ideally, such a sample would have a score of 0 since the technically correct answer is 100, not 120. However, the SFT $M_{Q\to A}$ student we use to construct a score for each $(Q, A)$ sample computes the final answer as 120 a staggering 47\% of the time. The verifier makes the exactly the same mistakes the $M_{A \to A}$ model made when constructing the question, likely because they were trained on similar distributions.

We suspect using a larger model more capable of detecting these sort of trivial non-solutions would do substantially better at generating backwards synthetic data. Similarly, employing separate models as the generator and verifier may reduce the probability of both making the same mistakes, improving the reliability of the score for each pair. We leave this as future work.

\section{RCRL Step-label Generating Process}
\label{sec:sorm}

Another natural candidate which could be used to identify mistakes at each step is a Process Based Reward Model (PRM) \citep{Lightman2023LetsVS}. A PRM estimates the probability of correctness of a step $S_i$, $p(S_i\texttt{ correct}|Q,S_1,S_2,...,S_i)$ independently of its impact on the final answer. However, this would be expensive, requiring collecting human annotated samples. Instead, we propose to approximate the \textit{optimal value function} $V^*$ of the reasoning task. $V^*$ corresponds to the value function of the \textit{optimal policy} which  is able to successfully solve the reasoning task from any logically valid intermediate state $S_j$. Such an optimal value function would have $V^*(Q, S_1,..., S_i) = 1$ for a solution prefix with no mistakes, and $V^*(Q, S_1, ..., S_i) = 0$ if the prefix already contains a mistake which will result in an incorrect final answer. Note however, $V^*$ does not exactly correspond to a PRM. This is because a partial solution $S_1,...,S_i$ with a mistake at step $j \neq i$ and valid terminal step $S_i$ will have $V^*(Q,S_1,...,S_i) = 0$ and $PRM(Q, S_1,...,S_i)=1$. To make this distinction clear, we call models we train to directly approximate $V^*$ stepwise ORMs or \textbf{SORMs}.

\end{document}